\documentclass[twoside,11pt]{article}

\usepackage{blindtext}
\usepackage{lipsum}
\usepackage{amsmath}
\usepackage{algorithm}
\usepackage{algorithmic}
\usepackage{booktabs}
\usepackage{multirow}
\usepackage{mathtools}
\usepackage[dvipsnames]{xcolor}
\usepackage{xurl}
\usepackage{subcaption}
\usepackage{jmlr2e}
\usepackage{amsmath}

\newcommand{\T}{\ensuremath{^\top}}

\renewcommand{\v}[1]{\mathbf{#1}}
\defcitealias{chizat2018:scaling}{C18}  

\usepackage{lastpage}
\jmlrheading{}{}{1-\pageref{LastPage}}{}{}{}{Tomasz Kacprzak, Michael W. Heiss, Gianluca Janka, Ann M. Dillner, Satoshi Takahama}
\ShortHeadings{Optimal Transport Linear Models}{Optimal Transport Linear Models}
\firstpageno{1}

\begin{document}

\title{Scalable Approximate Algorithms for Optimal Transport Linear Models}

\author{\name Tomasz Kacprzak \email tomasz.kacprzak@psi.ch \\
       \addr Swiss Data Science Center, Paul Scherrer Institute, 5232 Villigen, Switzerland
       \AND
       \name Francois Kamper \\
       \addr Swiss Data Science Center, Ecole Polytechnique Fédérale de Lausanne, 1015 Lausanne, Switzerland
       \AND
       \name Michael W. Heiss, Gianluca Janka\\
       \addr PSI Center for Neutron and Muon Sciences CNM, 5232 Villigen PSI, Switzerland
       \AND
       \name Ann M. Dillner \\
       \addr Air Quality Research Center, University of California, Davis, CA 95618, USA
       \AND
       \name Satoshi Takahama \\
       \addr Laboratory for Environmental Spectrochemistry, Ecole Polytechnique Fédérale de Lausanne, 1015 Lausanne, Switzerland
       }

\editor{My editor}

\maketitle

\begin{abstract}
  Recently, linear regression models incorporating an optimal transport (OT) loss have been explored for applications such as supervised unmixing of spectra, music transcription, and mass spectrometry.
  However, these task-specific approaches often do not generalize readily to a broader class of linear models.
  In this work, we propose a novel algorithmic framework for solving a general class of non-negative linear regression models with an entropy-regularized OT datafit term, based on Sinkhorn-like scaling iterations.
  Our framework accommodates convex penalty functions on the weights (e.g. squared-$\ell_2$ and $\ell_1$ norms), and admits additional convex loss terms between the transported marginal and target distribution (e.g. squared error or total variation).
  We derive simple multiplicative updates for common penalty and datafit terms.
  This method is suitable for large-scale problems due to its simplicity of implementation and straightforward parallelization.
\end{abstract}


\section{Introduction}

Wasserstein distances \citep{kantorovich1960:wasserstein}, Sinkhorn distances \citep{cuturi2013:sinkhorn} and other general optimal transport (OT) losses have been used in multiple applications, such as unsupervised domain adaptation \citep{frogner2015:domainadaptation}, training of machine learning models \citep{montesuma2024:otml}, generative modelling \citep{arjovsky2017:wgan}, continuous normalizing flows \citep{onken2021otflow}.
In balanced OT, the mass of the given source and target distributions match exactly, and the entire mass of the source is transported to the target.
The marginals of the transport plan corresponding to the source (target) are equal to the given source (target) distributions.
In unbalanced OT, the masses of the source and target distributions can differ, and the total transported mass is found as a part of the optimization problem.
The agreement between the transport marginals and the input distributions is encouraged using convex datafit terms added to the loss function.
Various datafit functions can be used, allowing for tailoring of the model to the problem at hand.

Several works have proposed using optimal transport distances for linear regression type problems for non-negative measures, beyond probability distributions.
We refer to these models as \emph{Optimal Transport Linear Models} (OTLM), encompassing optimal transport losses, as a part of the family of Generalized Linear Models (GLMs).
In OTLM, the source marginal is created by a non-negative linear combination of the dictionary basis vectors.
It is then pushed forward via a transport plan, creating the target marginal.
The target marginal is then compared to the given target distribution.
The basis weights are found jointly with the transport plan by minimizing the (regularized) transport loss.
The optimal transport loss is well suited for problems where the difference between the model and target data is preferred to be measured as perturbations of the ``shape'', rather than additive or shot noise per sample.
This is desirable in many applications, in particular when there are discrepancies between the model and the target data are due to calibration errors, systematic effects, or other non-random effects.
Such models have been proposed for several applications, including 
modeling of mass spectra \citep{ciach2024:masserstein}, 
music transcription \citep{flamary2016:music}, 
and unmixing of planetary spectra \citep{nakhostin2016:supervised}.

Solving OT problems with the use of entropy regularizer has been widely adapted due to its many attractive properties: 
uniqueness of the solution \citep{cuturi2013:sinkhorn}, 
the recovery of the unregularized OT plan for entropic regularization parameter $\epsilon\!\to\!0$ \citep{cominetti1994:asymptotic,weed2018:explicit}, and the existence of simple, efficient, scalable and parallelizable algorithms, based on the iterative alternating updates \citep{sinkhorn1964:relationship, cuturi2013:sinkhorn, peyre2019:computational, benamou2015:iterative}, with well understood convergence properties \citep{bernton2022:entropic}
Unbalanced optimal transport problems with entropic regularization can be efficiently solved using \emph{Scaling Algorithms} \citep[][herafter \citetalias{chizat2018:scaling}]{chizat2018:scaling}, which rely on the Dykstra-like alternating multiplicative updates of the dual variables.

\vspace{-0.05cm}
\paragraph{Contributions}
In this work, we propose a generalized formalizm for the OTLM problems with entropic regularization.
We consider two types of problems: balanced and unbalanced OTLM.
In both cases, the source marginal is equal to the source distribution, and is a non-negative linear combination of the given dictionary basis vectors.
In the balanced case, the target marginal and the target distribution match exactly. 
In the unbalanced case, they can differ, and the corresponding loss term is added to the optimization objective.
The updates for several common losses are known; we derive new updates for the squared error loss and the negative Poisson likelihood loss.
The weight penalty can be used with any strictly convex function.
Similarly to the classic OT problems, we approximate the solution to the unregularized problem by adding an entropic regularization term.
We show how to extend the scaling algorithm to the OTLM problems, including the weight penalties.
The proposed algorithm uses Sinkhorn-like alternating multiplicative updates: 
the target marginal is updated as in \citetalias{chizat2018:scaling}, 
while the dictionary weights are updated using majorization-minimization steps.
We derive the MM updates for the following penalties:
the squared-$\ell_2$ norm (corresponding to the Ridge regression problem),  
$\ell_1$ (LASSO), 
and their mix (Elastic Net).
This method is suitable for large problems due to its simplicity of implementation and straightforward parallelization.

\paragraph{Existing work}

The entropy-regularized balanced OT for unmixing was previously proposed by \citet{nakhostin2016:supervised} for application to analysis of planetary spectra.
The problem consisted of two Wasserstein loss terms: one for the target distribution and one for the weights prior.
\citet{flamary2016:music} proposed an OT approach for music transcription, where the use of the OT loss addressed the difficulty posed by shape perturbations of spectrum of individual notes, which is hard to tackle using the traditional dictionary learning approach.
The source consisted of a dictionary of delta functions at base harmonic frequencies, which simplified the optimization problem and allowed for a single-step solution.
OT regression for mass spectrometry was previously proposed by \citet{ciach2024:masserstein}, where an unregularized OT problem was cast as a linear program and solved using simplex methods.
It demonstrated that OT regression can recover molecule proportions accurately without extensive preprocessing of spectra required by other methods.

\paragraph{Notation}
We use $\v{1}$ to denote the vector of ones of the appropriate dimension. 
The symbols ``$\odot$'' and ``/'' are used to denote the element-wise product and division of two vectors or matrices, respectively.
The angle brackets $\langle \cdot, \cdot \rangle$ are used to denote the Frobenius inner product of two matrices.
Bold letters are used for vectors and matrices, and their corresponding elements with the same letter in lowercase.
We use the $\ell_1$ and $\ell_2$ norms defined as $\ell_1(\cdot)\!=\!\|\cdot\|_1$ and $\ell_2(\cdot)\!=\!\|\cdot\|_2$, respectively.
The squared error loss is defined as $\mathrm{L2}(\cdot | \cdot) = \frac{1}{2} \|\cdot\!-\!\cdot\|_2^2$, and the total variation (TV) loss is defined as $\mathrm{TV}(\cdot | \cdot) = \|\cdot\!-\!\cdot\|_1$.

\section{Background}

\subsection{Generalized Linear Models}

Generalized linear models (GLMs) are a popular class of models in modern machine learning and signal processing \citep{nelder1972glm}.
Classical examples include linear regression and logistic regression, the LASSO \citep{tibshirani1996:regression}, and the Elastic Net \citep{zou2005:elastic}.
GLMs find optimal weights of a given basis using an optimization problem consisting of two parts: the datafit term and the penalty term.
Most commonly used models are convex in both the datafit and penalty terms, but non-convex datafits or penalties can also be used.
In this work, we consider GLMs with a convex loss function $\mathrm{L}$ and convex penalty function $\mathrm{R}$, which can be written as:
\begin{align} \label{eq:glm}
    \min_{\v{w} \in \mathbb{R}^M} \quad& \mathrm{L}(\v{X} \v{w} | \v{b}) + \alpha \mathrm{R}(\v{w}).
\end{align}
The algorithms for solving GLMs are well studied and benefit from the extensive algorithmic and implementation work \citep{bertrand2022:skglm}.
Most of the GLM problems are formulated in a way that the loss is computed as a sum of per-sample losses, which poses difficulties when the dictionary basis vectors are not aligned well with the data in terms of their location and shape.

\subsection{Optimal Transport}

We first review the balanced optimal transport problem, and then extend it to the unbalanced version. 
In a discrete setting, the Kantorovich optimal transport problem with entropic regularization finds a transport plan $\v{Q} \in \mathbb{R}_{+}^{I \times J}$ that minimizes the cost of transporting the source distribution $\v{a} \in \mathbb{R}_{+}^I$ to the target distribution $\v{b} \in \mathbb{R}_{+}^J$, under a penalty on the transport plan.
Generally, the source and target distributions can be of different dimensionality.
The cost of transporting one unit of mass from the $i$-th source point to the $j$-th target point is given by $C_{ij}$.

\paragraph{Balanced OT}
In the balanced case, the source and target distributions match the source and target marginals, that is, $\v{a\T 1}\!=\!\v{b\T 1}$.
The balanced optimal transport problem is
\begin{equation}
  \min_{\v{Q} \in \Pi(\v{a}, \v{b})} \ \ \langle \v{C}, \v{Q} \rangle,     \tag{OT}
\end{equation}
where $\Pi(\v{a}, \v{b}) = \left\{ \v{Q} \in \mathbb{R}_{+}^{I \times J} \mid \v{Q} \v{1} = \v{a}, \ \v{Q}\T \v{1} = \v{b} \right\}$ is the set of all transport plans that satisfy the marginal constraints.
This is a linear program and its solution may not be unique.
The solution can be calculated using linear programming with $\mathcal{O}(n^3 \log(n))$ complexity, which can be difficult to scale to large problems.
By adding the a penalty on the transport plan, the problem becomes strictly convex and the solution is unique.
It can be solved using the celebrated Sinkhorn algorithm \citep{sinkhorn1964:relationship}, which is simple, efficient, parallelizable, and scales well to large problems.
The balanced optimal transport problem with entropic regularization is
\begin{equation}
  \min_{\v{Q} \in \Pi(\v{a}, \v{b})} \ \langle \v{C}, \v{Q} \rangle + \epsilon \mathrm{KL}(\v{Q}|| \v{K}), \tag{$\epsilon$OT}
\end{equation}
where $\v{K} = \exp({-\v{C}/\epsilon})$ is the Gibbs kernel, and KL is the Kullback-Leibler divergence:
\begin{equation}
  \label{eq:generalized_kl}
    \textstyle \mathrm{KL}(\v{X} || \v{Y}) = \sum_{i,j} X_{ij} \log (X_{ij}/Y_{ij}) - X_{ij} + Y_{ij},
\end{equation}
with convention \mbox{$0 \log 0 = 0$}, and $\epsilon\!>\!0$ is the regularization parameter.
The total transported mass is fixed to the total mass of the distributions.
In the limit of $\epsilon\!\to\!0$, the solution to the $\epsilon$OT problem converges to the solution of the OT problem.

\paragraph{Optimal transport distance}
Optimal transport cost is a distance whenever the matrix $\v{C}$ is itself a metric matrix \citep{villani2009:optimal}.
Optimal transport distances have gained a lot of attention in the machine learning community following the work of \citet{cuturi2013:sinkhorn}, as the entropic approximation benefits from the efficiency of the Sinkhorn algorithm and enables calculating this distance at scale.

\paragraph{Unbalanced OT}
In the unbalanced problem (UOT), deviations are allowed between the transport marginals and the input distributions, so that $\v{Q} \v{1} \!\neq\!\v{a}$ and $\v{Q}\T \v{1} \!\neq\!\v{b}$.
These deviations are penalized using convex and lower-semicontinuous functions $\mathrm{F}_{1} : \mathbb{R}_{+}^{I} \to \mathbb{R}_{+} \cup \{\infty\}$ and $\mathrm{F}_{2} : \mathbb{R}_{+}^{J} \to \mathbb{R}_{+} \cup \{\infty\}$:
\begin{equation*}
  \min_{\v{Q} \in \mathbb{R}^{I\!\times\!J}} \ \ \langle \v{C}, \v{Q} \rangle + \epsilon \mathrm{KL}(\v{Q}|| \v{K}) + \mathrm{F}_{1}(\v{Q} \v{1}) + \mathrm{F}_{2}(\v{Q}\T \v{1}) .
\end{equation*}
Typically, the functions $\mathrm{F}_1$ and $\mathrm{F}_2$ penalize the deviations from the source $\v{a}$ and target $\v{b}$ marginals, respectively.
The commonly used functions include the Kullback-Leibler divergence or the total variation distance.
Balanced optimal transport can be viewed as a special case of unbalanced optimal transport, where the marginals match the source and target distributions exactly: the function $\mathrm{F}$ becomes an indicator function 
$\mathrm{F}_1(\cdot)\!=\!\iota_{\{=\}}( \cdot | \v{a})$ and 
$\mathrm{F}_2(\cdot)\!=\!\iota_{\{=\}}( \cdot | \v{b})$, where $\iota_{\{=\}}( \cdot | \cdot)$ is the indicator function, which is 0 if the arguments are equal and $\infty$ otherwise.
More generally, a sum of any number of functions can be used that encode soft or hard constraints on the transport plan \citepalias{chizat2018:scaling}.

\subsection{Scaling Algorithms for Optimal Transport} \label{sect::scale_algs}
There exist several algorithms for UOT \citep{sejourne2023:unbalanced}.
This work builds on the \emph{Scaling Algorithm} \citepalias{chizat2018:scaling}, a generalization of the Sinkhorn algorithm.
The scaling algorithm uses updates alternating between the source and target marginals. 
The updates are multiplicative, which ensures the non-negativity of the transport plan.
As shown in \citetalias{chizat2018:scaling}, the scaling algorithm converges if $\mathrm{F}_1 , \mathrm{F}_2$ are a sum of convex lower-semicontinuous functions. For such a function $\mathrm{F}$, the alternating updates are calculated using a proximal operator defined by:
\begin{equation}
  \label{eqn:prox}
    \mathrm{prox}^{\mathrm{KL}}_{\mathrm{F}}(\v{y}) = \arg \min_{\v{s}} \mathrm{F}(\v{s}) + \epsilon\mathrm{KL}(\v{s} || \v{y}).
\end{equation}
The functions $\mathrm{F}$ can be defined through an auxiliary minimization problem and can still be handled efficiently by the scaling algorithm \citepalias{chizat2018:scaling}. 
For the UOT problems with two marginal constraints, the updates are given by:
\begin{equation}
    \v{u}_2 \leftarrow \frac{\mathrm{prox}^{\mathrm{KL}}_{\mathrm{F}_1}(\v{Ku}_1)}{\v{Ku}_1},     \quad
    \v{u}_1 \leftarrow \frac{\mathrm{prox}^{\mathrm{KL}}_{\mathrm{F}_2}(\v{K\T u}_2)}{\v{K\T u}_2},     
\end{equation}
starting from $\v{u}_1\!=\!\v{1}$.
The final transport plan is given by $\v{Q} \leftarrow \v{K} \odot \v{u}_1 \v{u}_2\T$.
For the balanced case where a hard constraint is placed on the marginals, the updates are given by:
    \mbox{$\v{u}_2 \leftarrow \v{a} / \v{Ku}_1$}, \mbox{$\v{u}_1 \leftarrow \v{b} / \v{K\T u}_2$}, 
which is similar to the original Sinkhorn algorithm.
For a number of common functions, the proximal operator can be computed using a simple update, see Table~\ref{tab:prox}.
This alternating scheme can be used for UOT variants with any number of soft or hard marginal constraints.

\section{Optimal Transport Linear Models}

We propose a formalization of the regularized OTLM problem.
The source marginal is interpreted as a linear combination of a non-negative dictionary basis $\v{X} \in \mathbb{R}^{N \times M}_+$ with weights $\v{w} \in \mathbb{R}^M_+$. 
This way, the source marginal is constrained to lie in the non-negative span of the dictionary basis. 
The target distribution is the given data vector $\v{y} \in \mathbb{R}^N_+$. 
The matrix $\mathbf{C} \in \mathbb{R}^{N \times N}_+$ can be specified using prior knowledge about the problem.
This formulation allows to encode the OTLM problem in a way that admits the efficient scaling algorithm introduced in Section \ref{sect::scale_algs}.

To ensure that the source marginal of the transport plan match the linear combination of weights and basis exactly, $\v{X} \v{w}\!=\!\v{Q} \v{1}$, and to include a penalty on the weights, we propose to define $\mathrm{F}_{1}(\v{s})\!=\!\alpha \mathrm{P_R}(\v{s}| \v{X})$ through the following auxiliary convex problem
\begin{equation}
  \label{eq:P}
\mathrm{P_R}(\v{s}| \v{X}) \;=\; 
\inf_{\v{w} \in \mathbb{R}^M} \left\{  \mathrm{R}(\v{w}) + \iota_{\{=\}} (\v{X} \v{w} | \v{s}) \right\},
\end{equation}
where the convex function $\mathrm{R}(\cdot)$ represents a penalty term, analogous to the penalty function in GLMs (see the following section), and $\alpha$ is a scalar regularization parameter.
Thus, $\mathrm{P_R}(\v{s}| \v{X})$ is the minimal distance (determined by R) from the origin to the subspace defined by $\v{Xw}\!=\!\v{s}$.
By setting $\mathrm{F}_{2}(\boldsymbol{s})\!=\!\lambda \mathrm{L}(\boldsymbol{s}|\boldsymbol{y})$, with $\mathrm{L}(\cdot|\cdot)$ suitably chosen to satisfy the UOT conditions in the first argument, we encourage the column (target) marginals 
of $\v{Q}$ to be close to the target distribution. 
This function is chosen to reflect the type of the 
regression problem, for example $\mathrm{L}(\boldsymbol{s}|\boldsymbol{y})\!=\!\frac{1}{2} \|\boldsymbol{s}-\boldsymbol{y} \|_{2}^{2}$ in the case of continuous non-negative data, and $\lambda\!\geq\!0$ is a corresponding scalar regularization parameter.
With scalar regularization parameter $\alpha\!\geq\!0$, the problem can be written as: 
\begin{flalign}
  \begin{aligned}
    \min_{\v{Q} \in \mathbb{R}^{N\!\times\!N}_+} \ \langle &\v{C}, \v{Q} \rangle + \epsilon \mathrm{KL}(\v{Q} || \v{K})  + \alpha \mathrm{P_R}(\v{Q} \v{1}| \v{X}) + \lambda \mathrm{L} (\v{Q}\T \v{1} | \v{y}).
\end{aligned}
\tag{OTLM}\label{eq:otlm}
\end{flalign}    
which is a form admitting the scaling algorithm, or alternatively,
\begin{equation}
  \min_{\v{Q} \in \mathbb{R}^{N\!\times\!N}_+, \v{w} \in \mathbb{R}^M_+} \ \langle \v{C}, \v{Q} \rangle + \epsilon \mathrm{KL}(\v{Q} || \v{K})  + \lambda \mathrm{L} (\v{Q}\T \v{1} | \v{y}) + \alpha \mathrm{R}(\v{w}), \quad \text{s.t.} \quad \v{Q} \v{1} = \v{X} \v{w},
\end{equation}
which highlights the connection between the \ref{eq:otlm} and the GLM objective in Equation~\ref{eq:glm}, as the same functions for datafit L and penalty R can be used.
From now we will drop the subscript R in $\mathrm{P_R}$ for brevity.

\section{Scaling Algorithm Framework for OTLM}

In order to solve the OTLM problem in the scaling algorithm framework, we need to calculate proximal operators for $\mathrm{L}$ and $\mathrm{P}$, and then follow the iterations derived in \citetalias{chizat2018:scaling}.
The proximal operators for various functions $\mathrm{L}$ have been derived in the same work. 
We additionally derive the proximal operator for the L2 loss and the negative Poisson likelihood, and gather them in Table~\ref{tab:prox}.
Note that the negative Poisson likelihood is a special case of the KL divergence in reverse mode.
See Appendix~\ref{app:l2marginal} for derivations.
Substituting Equation~\ref{eq:P} into the definition in Equation~\ref{eqn:prox}, we obtain the proximal operator for $\alpha \mathrm{P}$:
\begin{equation*}
     \mathrm{prox}^{\mathrm{KL}}_{\alpha \mathrm{P}}(\v{u}) = \v{Xw},  \ \  \v{w} = \arg \min_{\v{w} \geq \v{0}}  \ \alpha \mathrm{R}(\v{w}) + \epsilon \mathrm{KL}(\v{Xw} || \v{u}). \label{eqn:prox_nnsx} \tag{ProxP} 
 \end{equation*}
The form of this proximal operator draws a further connection between OTLM and the GLM objective. At convergence the 
value of $\v{w}$ is be determined by 
$\v{w} = \arg \min_{\v{w} \geq \v{0}}  \ \alpha \mathrm{R}(\v{w}) + \epsilon \mathrm{KL}(\v{Xw} || \v{K} \v{u}_{1})$ which differs from (\ref{eq:glm}) by exchanging  $\mathrm{L}(\v{Xw} | \v{y})$ with $\epsilon \mathrm{KL}(\v{Xw} || \v{K} \v{u}_{1})$. Since at convergence we have $\v{K} \v{u}_{1} = \v{Q}^{\top} \v{1} / \v{u}_2 \approx \v{y} / \v{u}_2$ we can interpret our regression coefficients as being determined by comparing $\v{Xw}$ to a rescaling of 
the target values via the KL divergence, subject 
to a penalty.  

The optimization problem in \ref{eqn:prox_nnsx} does not have a closed-form solution and needs to be evaluated numerically. 
We propose to solve this problem using an iterative majorization-minimization (MM) scheme, where the objective function is majorized using Jensen's inequality. 
If this evaluation can be done to a sufficient numerical accuracy, we can rely on the convergence results of \citetalias{chizat2018:scaling}. 

\subsection{MM for Penalized Column Span Projection with KL Divergence}

The objective of the optimization problem in \ref{eqn:prox_nnsx} is:
\small
\begin{equation*}
  \mathrm{V}(\v{w})\!=\!\alpha \mathrm{R}(\v{w})\!+\!\epsilon\left( \sum_{i=1}^N  ( \v{Xw})_i \log\frac{(\v{Xw})_i}{y_i}\!-\!(\v{Xw})_i\!+\!y_i \right).
   \label{eqn:obj_nnsx}
\end{equation*}
\normalsize
The KL divergence term contains a sum over elements of $\v{w}$, which prohibits a direct solution in a closed form.
We propose the following majorization-minimization scheme, where the objective function is majorized using the Jensen's inequality.
We use the type of majorization used frequently in the non-negative matrix factorization (NNMF) algorithms \citep{lee2000:nnmf}.
\begin{table}
  \vspace{-0.5cm}
  \centering
  \renewcommand{\arraystretch}{1.2} 
  \begin{tabular}{p{3cm} p{10cm}}
  \toprule
  Datafit $\mathrm{L}$          & $\mathrm{prox}^{\mathrm{KL}}_{\mathrm{L}}(\mathbf{s}| \v{y})$ \\ \midrule
  \small \(\iota_{\{=\}}(\mathbf{s} | \mathbf{y})\)           & \small \(\mathbf{y}\) \\
  \small \(\lambda \mathrm{KL}(\mathbf{s} | \mathbf{y})\)     & \small \(\mathbf{s}^{\frac{\epsilon}{\epsilon + \lambda}} \cdot \mathbf{y}^{\frac{\lambda}{\epsilon + \lambda}}\) \\
  \small \(\lambda \mathrm{TV}(\mathbf{s} | \mathbf{y})\)     & \small \(\min \{\mathbf{s} \cdot \exp( \frac{\lambda}{\epsilon}), \max \{\mathbf{s} \cdot \exp(-\frac{\lambda}{\epsilon}), \mathbf{y}\}\}\) \\
  \small \(\lambda \mathrm{L2}(\mathbf{s} | \mathbf{y})\)        &  \small $\frac{\varepsilon}{\lambda} \mathrm{W}_0 \left(\frac{\lambda}{\varepsilon} \mathbf{s} \exp\left(\frac{\lambda}{\varepsilon} \mathbf{y}\right)\right).$ \\
  \small \(\lambda \mathrm{Poiss}(\mathbf{s} | \mathbf{y})\) & \small 
$
\frac{\lambda}{\varepsilon} \v{y}
\;
  \mathrm{W}^{-1}_0\! \left(
     \frac{\lambda\,\v{y}}{\varepsilon\,\v{s}}
     \,\exp \left( \frac{\lambda}{\varepsilon} \right)
  \right)
$
\\ \bottomrule
  \end{tabular}
  \caption{Marginal loss functions and their associated $\mathrm{prox}_{\mathrm{L}}^{\mathrm{KL}}$ operators, following \citetalias{chizat2018:scaling}.
  $\mathrm{Poiss}$ is the negative Poisson log-likelihood. 
  $\mathrm{W}_0$ is the principal branch of the Lambert W function.
  The last two functions were added in this work.
  } 
  \label{tab:prox}
\end{table}
\begin{table}
  \centering
  \renewcommand{\arraystretch}{1.5} 
  \begin{tabular}{p{3cm} p{10cm}}
    \toprule
  Penalty R          & $\mathrm{wMMstep}^{\mathrm{KL}}_{\mathrm{P}}(\mathbf{y}, \v{Z} | \v{X})$ \\ \midrule
  \small None &
  \small $w_j \!\leftarrow\!\exp \left( \frac{1}{\sum_i X_{ij}} X_{ij} \log \left( \frac{Z_{ij} y_i}{X_{ij}} \right) \right)  $ \\
  %
  \small $\alpha \ell_1 (\v{w})$ &
  \small $w_j \!\leftarrow\!\exp \left( \frac{1}{\sum_i X_{ij}} \sum_i X_{ij} \log \left( \frac{Z_{ij} y_i}{X_{ij}} \right)\!-\!\frac{\alpha}{\epsilon} \right)  $ \\ 
  \small $ \frac{\alpha}{2} \ell_2^2 (\v{w})$ &
  \small $w_j \!\leftarrow\!\frac{\epsilon}{\alpha} \sum_i X_{ij} \mathrm{W}_0 \left( \frac{\alpha}{\epsilon} \frac{1}{\sum_i X_{ij}} \exp \left( - \frac{X_{ij}}{Z_{ij} y_i} \right) \right) $ \\
  \small $\alpha \ell_1(\v{w})\!+\!\frac{\beta}{2} \ell_2^2(\v{w})$ &
  \small $w_j\!\leftarrow\!\frac{\alpha}{\beta} \mathrm{W}_0 \left( \frac{\beta}{\alpha}\frac{1}{\sum_i X_{ij}} \exp\left(  -\frac{1}{\sum_i X_{ij}} \left( \frac{\alpha}{\epsilon}\!+\! \sum_{i=1}^{N} X_{ij} \log\left( \frac{X_{ij}}{Z_{ij} y_j} \right) \right) \right) \right)$
\\ 
\bottomrule
  \end{tabular}
  \caption{MM algorithm steps for the proximal operators for projection into non-negative column span of matrix $\v{X}$ under different penalties $\mathrm{R}$.
  The variable $x_j\!\coloneqq\!\sum_i X_{ij}$ is the sum of the $j$-th column of $\v{X}$, and 
  $Z_{ij}\!\coloneqq\!X_{ij} w'_{j} / (\sum_{k=1}^M X_{ik} w'_{k})$ is the matrix of the normalized weights $w'$ computed from the previous iteration.
  }
  \label{tab:prox_nnsx}
\end{table}
\begin{algorithm}
  \small
  \begin{algorithmic}
  \REQUIRE Cost matrix $\v{C}$, feature matrix $\v{X}$, target data $\v{y}$ \vskip 1pt
  \REQUIRE Regularization parameters $\epsilon$, $\alpha$, $\lambda$, initial weights $\v{w}$ \vskip 1pt    
  \STATE Initialize $\v{K} = \exp(-\v{C}/\epsilon)$, $\v{u}_1 = \v{1}$ \vskip 1pt
  \WHILE{not converged} \vskip 1pt
     \STATE $\triangleright$ \textit{Source update: Penalized non-neg. column span projection proximal step} \vskip 1pt
     \STATE $Z_{ij} \leftarrow X_{ij} w_{j} / \sum_{k=1}^M X_{ik} w_{k}  \quad \forall \ i,j$ \vskip 1pt
     \STATE $\v{w} \leftarrow \mathrm{wMMstep}^{\mathrm{KL}}_{\mathrm{P}}(\v{Ku}_1, \v{Z}| \v{X})$ \quad $\triangleright$ \textit{single MM step}  \vskip 1pt 
     \STATE $\v{u}_2  \leftarrow \v{Xw} / \v{Ku}_1$ \vskip 1pt
     \STATE $\triangleright$ \textit{Target update: datafit proximal step} \vskip 1pt
      \STATE $\v{v}_2 \leftarrow \mathrm{prox}^{\mathrm{KL}}_{\mathrm{L}}(\v{K}^\top\v{u}_2| \v{y})$  \vskip 1pt
      \STATE $\v{u}_1 \leftarrow \v{v}_2 / \v{K}^\top \v{u}_2$ \vskip 1pt
  \ENDWHILE \vskip 1pt
  \STATE $\v{Q} \leftarrow \v{K} \odot \v{u}_1 \v{u}_2\T$ \vskip 1pt
  \STATE \textbf{return} $\v{w}$, $\v{Q}$
  \end{algorithmic}
  \caption{Scaling Algorithm for OTLM}
  \label{alg:otlm}
  \end{algorithm}
Similar technique was recently applied for solving unregularized unbalanced OT by \citet{chapel2021:unbalanced}.
Define the matrix $\v{Z}\!\in\!\mathbb{R}^{N \times M}_+$ with elements $Z_{ij}\!=\!X_{ij} w'_{j} / \sum_{k=1}^M X_{ik} w'_{k}$, where $w'_{k}$ is taken from the previous iteration.
The majorized objective is
\begin{equation}
\begin{aligned}
\label{eqn:obj_nnsx_majorized}
\mathrm{G}(\v{w}, \v{w}')\!=\!\epsilon \sum_{i=1}^N \sum_{i=1}^M Z_{ij} \Biggl( \frac{X_{ij} {w}_{j}}{Z_{ij}} \log \left( \frac{ X_{ij} w_j}{ Z_{ij}} \frac{1}{y_i} \right)   
-\!\frac{ X_{ij} w_j}{Z_{ij}}\!+\!y_i \Biggr)\!+\!\alpha \mathrm{R}(\v{w}). 
\end{aligned}
\end{equation}
The function $\mathrm{G}(\v{w}, \v{w}')$ is convex in $\v{w}$ and $\v{w}'$, and majorizes the objective: $\mathrm{G}(\v{w}, \v{w}')\! \geq\!\mathrm{V}(\v{w})$, $\mathrm{G}(\v{w}, \v{w})\!=\!\mathrm{V}(\v{w})$, for any $\v{w}\!\geq\!0$.
A single update of the weights $\v{w}$ is:
\begin{equation*}  
    \v{w} \leftarrow \mathrm{wMMstep}^{\mathrm{KL}}_{\mathrm{P}}(\v{y}, \v{Z} | \v{X}) 
    = \arg \min_{\v{w} \geq 0} \mathrm{G}(\v{w}, \v{w}' | \v{y}, \v{X}).
\end{equation*}
We show the updates for the $\ell_1$, the squared-$\ell_2$, and the mixed $\ell_1 / \ell_2^2$ penalties, as well as for the unpenalized case, in Table~\ref{tab:prox_nnsx}.
Appendix~\ref{app:prox_nnsx} gives the derivations of these updates and the MM algorithm for the proximal step $\mathrm{Prox}_{\alpha \mathrm{P}}^{\mathrm{KL}}(\v{y}| \v{X})$.

\subsection{Iterative Algorithm for OTLM}
The final scaling algorithm for the OTLM problem is given by Algorithm~\ref{alg:otlm}.
In this algorithm, we update the weights $\v{w}$ using a single step of the MM algorithm.
In general, the MM algorithm for $\mathrm{prox}_{\alpha \mathrm{P}}^{\mathrm{KL}}(\v{Ku}_1| \v{X})$ could be run until convergence, but given that a single step guarantees a decrease in the objective, we find it more efficient to update the target after each single MM step.
Limiting the number of evaluations of a numerical procedure inside a loop has proven beneficial in other applications, see for example \citet{johnson2009fixing}. 
In our experiments we also observed faster convergence for a single MM step, but the benefit will be dependent on the problem at hand.
The computational complexity of this method is that of the Sinkhorn algorithm, which is typically $\mathcal{O}(n^2 \log(1/\epsilon))$.
  
\section{Demonstration Examples}

\begin{figure*}
  \hspace{0.5cm}
  \includegraphics[width=0.3\linewidth]{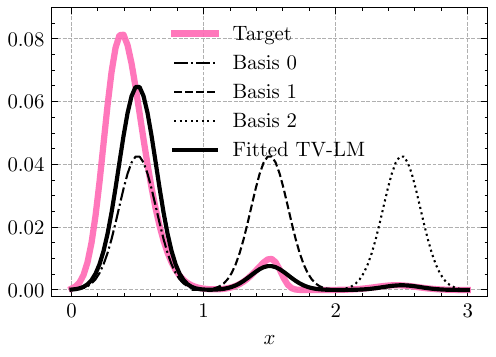}
  \includegraphics[width=0.3\linewidth]{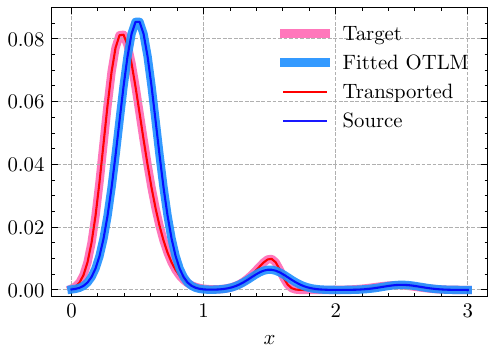}\hspace{1.2cm}
  \includegraphics[width=0.229\linewidth]{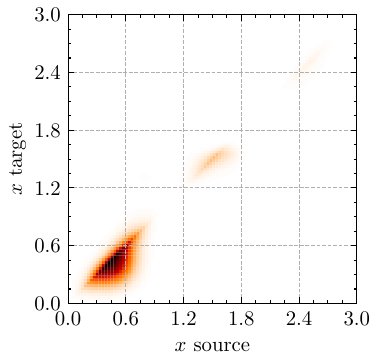}
  \vspace{-0.5cm}
  \caption{TV regression compared to the balanced OTLM, without weight penalties.
  Left: target data, three basis vectors, and the best fit using TV regression.
  Middle: balanced OTLM with the same basis vectors. 
  Right: the OTLM transport plan.
  }
  \label{fig:demo}
\hspace{0.5cm}
  \includegraphics[width=0.3\linewidth]{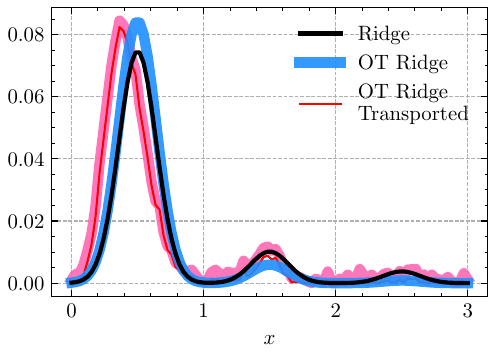}
  \includegraphics[width=0.3\linewidth]{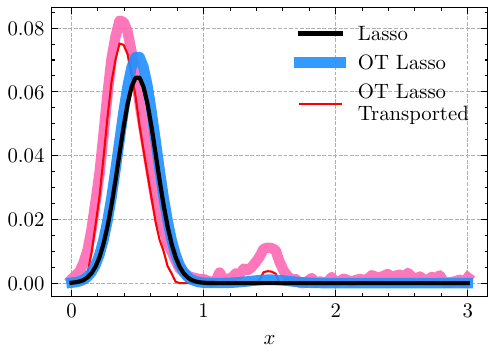}
  \includegraphics[width=0.3\linewidth]{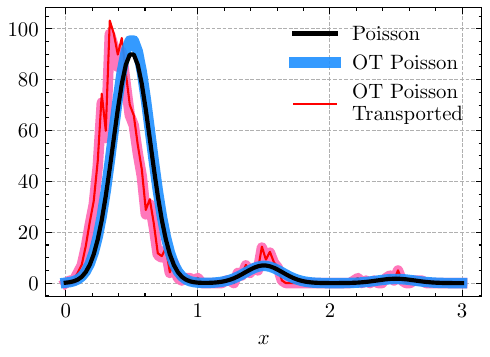}
  \vspace{-0.5cm}
  \caption{Example solutions for Ridge, LASSO, and Poisson regression (black) and their OTLM variants (blue).}
  \label{fig:demo_multiple}
\end{figure*}

We demonstrate the solutions of the OTLM problem on a simple simulated dataset resembling spectroscopic data.
We consider a problem of fitting the weighed basis vectors to the target distribution. 
The target data $\v{y}$ is a linear combination of the biased basis vectors, which we consider to be unknown.
We are given a matrix $\v{X}$ of unbiased basis vectors.
We consider both the biased and unbiased basis vectors to sum up to 1.
The goal is to find the weights closest to the true weights of the biased basis vectors. 
The three basis vectors and the target distribution are shown in the left panel of Figure~\ref{fig:demo} with dashed black lines and the thick pink line, respectively.
The basis vectors are generated using Gaussian functions, with their means and variances chosen to be slightly different from the target data.
The target data is generated using three skew-Gaussian functions.
See Appendix~\ref{app:demo} for more details of construction of the demonstration problem, the formulation of the LM problems, and details of the solvers used to obtain their solutions.
The black line shows the solutions to the TV regression problem without the OT loss term.
The middle panel shows the solutions to the balanced OTLM problem, with 
the target distribution $\v{y}$ in the thick pink line, 
the target marginal $\v{Q\T1}$ in the thin red line (they are identical in this problem), 
the prediction $\v{Xw}$ in the thick light-blue line, 
and the source marginal $\v{Q1}$ in the thin blue line.
For a fully-converged solution, the source distribution $\v{Xw}$ and source marginal $\v{Q1}$ should be equal in OTLM, which is indeed the case here.
The right panel shows the corresponding optimal transport plan.
The values of the weight $\v{w}$ obtained with TV have lower values compared to the true weights of the basis vectors due to the misalignment of the basis.
OTLM is robust to this bias and recovers weights that are much closer to the true mass of the peak.

The unbalanced OTLM problem is demonstrated in Figure~\ref{fig:demo_multiple}.
The columns show the solutions to the classical problems: Ridge (left), LASSO (middle), and Poisson (right).
The pink lines show the target distributions.
The black lines correspond to the problems without the OT loss term.
These problems pose a challenge for the classical linear models; the use of ``biased'' basis vectors results in underestimation of their weights.
The blue lines show the solutions to the corresponding OTLM problems.
See Appendix~\ref{app:demo} for more details.
As in the balanced case, the OTLM solutions are closer to the true mass of the peaks.

\section{Scaling Examples}

\begin{figure}
  \centering
  \includegraphics[height=0.3\textwidth]{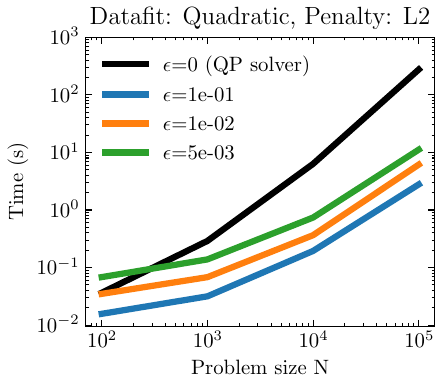}\hspace{-0.2cm}
  \includegraphics[height=0.3\textwidth]{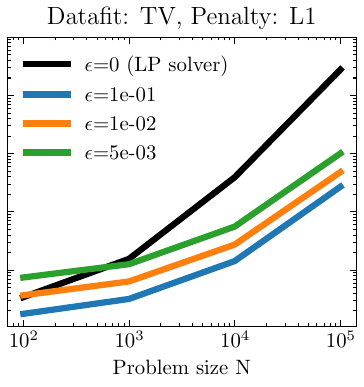}\hspace{0.5cm}
  \includegraphics[height=0.3\textwidth]{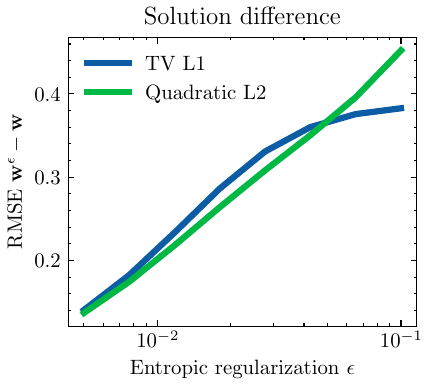}
  \vspace{-0.5cm}
  \caption{Scaling of the regularized OTLM algorithm compared to unregularized OTLM solved using linear (LP, left) and quadratic (QP, middle) programming solvers. 
  The right panel shows the difference between the weights obtained with $\epsilon$-regularized and unregularized OTLM.}
  \label{fig:scaling}
\end{figure}

Unregularized OTLMs can be solved using generic solvers, such as linear programming (LP), quadratic programming (QP), or other gradient-based methods, depending the choice of loss and penalty functions.
However, these solvers are computationally expensive for large problems.
Similarly to the classic OT  \citep{cuturi2013:sinkhorn}, the use of the entropic regularizer in OTLM allows to solve the problem approximately, but in much more efficient way.
We demonstrate the scalability of the proposed Algorithm~\ref{alg:otlm} to large problems.
We solve the unregularized and regularized OTLMs using the generic solvers and the proposed algorithm, respectively.
We compare their time-to-solution and the obtained values of the weights.
We consider two OTLM problems:
 (i) OT ridge regression with L2 loss and $\ell_2^2$ weight penalty,
(ii) OT total variation with the TV loss and $\ell_1$ penalty.
See Appendix~\ref{app:scaling} for the details of the formulation of the problems.
For the unregularized problems, we use the QP and LP solvers, respectively.
The test problems are created on a grid of $x$-coordinates, $\v{x} \in \mathbb{R}^{N}$ with $N$ ranging from $10^2$ to $10^5$.
The dictionary matrix $\v{X} \in \mathbb{R}^{N \times M}$ has $M=N/10$ basis vectors.
The basis vectors are Gaussian functions sampled at the $x$-coordinate.
The target $\v{y} \in \mathbb{R}^{N}$ is created as a mixture of components corresponding to the basis vectors, but with additional random perturbations in means and variances on the level of 20\%, and with additional skew drawn from the range of $[-2, 2]$.
The amplitudes of the mixture are chosen at random.
We used sparse representations of $\v{X}$ and $\v{C}$, with typically less than 10\% of non-zero elements.
See Appendix~\ref{app:scaling} for the details of the test problem generation and the configuration of solvers.
The experiment is repeated 32 times for each value of $N$.
The left and middle panels on figure Figure~\ref{fig:scaling} show the execution time of the unregularized and regularized OTLM, while the right panel shows the difference between the weights obtained with varying regularization strength $\epsilon$ compared to the unregularized OTLM.
The regularized OTLM solved using the scaling algorithm converges an order of magnitude faster than the equivalent solution obtained by solvers.
Notably, the memory footprint of the scaling algorithm is a small fraction of the memory footprint of the solvers.
The right panels show the difference between the weights $\v{w}^{\epsilon}$ obtained with varying regularization strength $\epsilon$ and weights $\v{w}$ obtained with the unregularized OTLM.
The approximation errors is calculated as the RMSE between the weights obtained with the unregularized and regularized OTLM.

\section{Results on datasets}

We compare the performance of the OTLM with the standard linear models on the following datasets: 
(i) muonic X-ray spectra of metal alloy standards, 
(ii) infrared spectroscopy of ambient particulate matter.

\subsection{Muonic X-ray spectra}
\label{sec:muonic_xray_spec}

\begin{figure}
  \centering
  \includegraphics[width=0.49\textwidth]{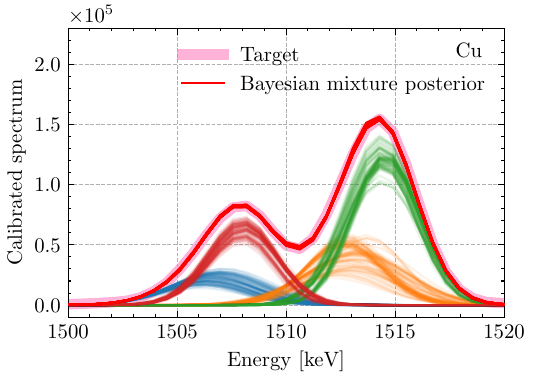}
  \includegraphics[width=0.49\textwidth]{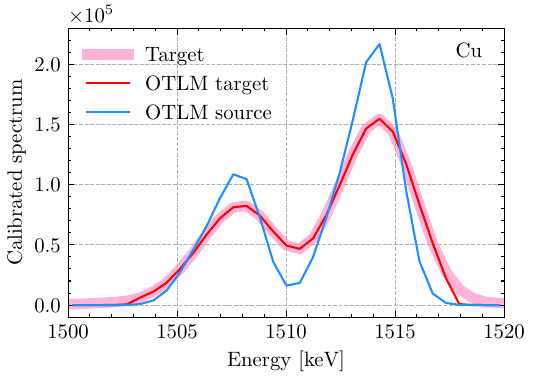}
  \caption{Spectrum of the BCR-691-A sample in the region of Cu $K_{\alpha}$ lines. Left: fit with non-convex Bayesian mixture model of skew-Gaussian distributions. Right: fit with convex OTLM. Both fits use 4 components.}
  \label{fig:muonic_xray_spec_data}
\end{figure}

\begin{figure}
  \centering
  \includegraphics[width=0.96\textwidth]{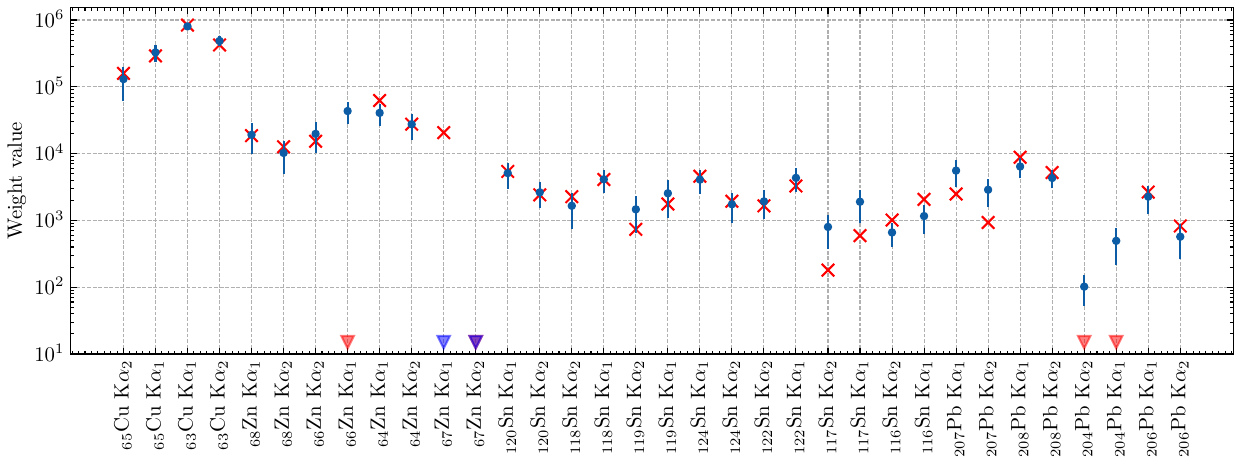}\hspace{0.1cm}
  \vspace{-0.3cm}
  \caption{Weights for each X-ray line of the BCR-691-A sample, as calculated by the convex OTLM method (red crosses) and non-convex mixture model of skew-Gaussian distributions (see Section~\ref{sec:muonic_xray_spec} for details).
  The mixture model weights were obtained using Baysian sampling.
  The error-bars show the standard deviation of the weights from the Bayesian posterior.
  The triangles show the OTLM weights that are outside the y-axis range.
   } \vspace{-0.8cm}
  \label{fig:muonic_xray_spec}
\end{figure}

Muonic X-ray spectra are used for elemental analysis \citep[see][for more information on these topics]{sturniolo2021:mudirac}.  
The goal of elemental analysis is to determine the relative weight fractions of the elements in the sample.
The weight fractions are proportional to the number of $K_{\alpha}$ transitions events and other calibration factors.
The measurement of the relative number of events between $K_{\alpha}$ transitions of different elements is therefore one of the key steps in the analysis.
Typically, the spectra are fitted with a mixture model by solving a non-convex optimization problem \citep{newville2016lmfit, hahn2024spectrafit}, which suffers from the following issues: multiple local minima, slow convergence, sensitivity to the initialization, and difficulty with fitting large number of components.
We show how OTLM can be used to fit the spectra very fast, with a unique solution.
In addition, OTLM scales linearly with the number of components, which allows for easy fitting of large dictionaries.
We compare the Bayesian mixture model fits with the OTLM fits, and show that OTLM recovers similar weights, in most cases consistent with the Bayesian posterior.
The spectrum data comes from measurements at the muon-induced X-ray emission (MIXE) beamline at the Paul Scherrer Institute \citep{gerchow:2022mixe}.
The samples used were the BCR-691 \citep{geel:bcr961} copper alloy standards.
The spectrum histogram was created from an event list and pre-processed to model the background.
Appendix~\ref{app:muonic_xray_spec} contains details of the data pre-processing, the Bayesian mixture model fitting and the OTLM fitting.

The reference method is a Bayesian mixture model of skew-Gaussian distributions.
Each component is parameterized by four parameters: the mean, standard deviation, skewness, and weight.
We describe the case for the BCR-691-A sample, which contains Cu, Zn, Sn, and Pb; other samples were analyzed in a similar manner.
The model has 36 components, corresponding to the $K_{\alpha 1}$ and $K_{\alpha 2}$ lines for the 4 elements and their isotopes.
The total number of parameters is 144.
Broad uniform priors on the parameters are used.
For the mean parameter, a small interval around the theoretical value was used to ensure identifiability of components.
The theoretical prediction for the X-ray energies is taken as predicted by the \textsc{Mudirac} code \citep{sturniolo2021:mudirac}.
Finally, the estimated background is added to the model.
We use the Poisson likelihood function.
We ran a sampler \textsc{Emcee} \citep{foreman2013emcee} with 500 walkers and 5000 steps, which took $\sim 10$ minutes on a single core of a 2021 MacBook Pro.
The results for the Cu $K_{\alpha}$ lines for the BCR-691-A sample are shown in the left panel of Fig.~\ref{fig:muonic_xray_spec_data}.
For OTLM, we create a dictionary of basis vectors $\v{X}\!\in\!\mathbb{R}^{N \times 34}$, corresponding to the same components.
We use Gaussian functions with means centered at the energies of these lines, as predicted by the \textsc{Mudirac} code \citep{sturniolo2021:mudirac}, and with the standard deviations taken from calibration measurements of the MIXE \citep{gerchow:2022mixe}.
The standard deviation of the peaks, as calculated previously from the calibration measurements, is somewhat mismatched for this sample; 
adjusting it would require additional non-convex procedure based on peak identification and spread measurements.
OTLM is robust to the shape and alignment mismatch of the peaks, which eliminates the necessity for such additional procedure.
The OTLM fit took $\lesssim 1$ seconds on the same machine and is shown in the right panel of Fig.~\ref{fig:muonic_xray_spec_data}.

The weight fractions calculated by both methods are shown in Fig.~\ref{fig:muonic_xray_spec}.
Some lines were measured close to zero, which are marked with triangles.
The OTLM fit is consistent with the Bayesian posterior for most of the lines.
As they are based on different models, the methods should not be compared directly and small differences are expected.
The mean absolute difference of the OTLM fit weights compared to the mean of the Bayesian posterior for all five BCR-691 samples and all elements is 0.83$\sigma$, where $\sigma$ is the standard deviation of the weights from the Bayesian posterior.
See Appendix~\ref{app:muonic_xray_spec} for details.

\subsection{Infrared Spectroscopy} \label{sect::IRS}
We consider an application using infrared spectroscopy to quantify the amount of airborne particles deposited on an air filter.  
A spectrum in this context measures absorbance, a ratiometric quantity computed from beam intensities with and without the sample present,  as a function of vacuum wavenumber, which is the inverse of the wavelength of electromagnetic radiation. 
The absorbance comprises additive contributions  from the collected particles as well as the collection media (i.e., the air filter). 

Here we consider spectra of several laboratory-generated analytes collected on polytetrafluoroethylene (PTFE) filters \citep{Ruthenburg2014}, namely 12-Tricosanone, ammonium sulfate, fructose, levoglucosan and suberic acid particles. 
Our objective is to estimate the amount of each compound present in each of the samples. To remove interference from the PTFE filters, we have 54 available spectra of blank filters without any analytes present.  
Furthermore, we have a spectral profile of a quantity proportional to the mass absorption coefficient (MAC) of pure particles for each analyte estimated from independent experiments \citep{Earle2006,SpectraBase}. 
Note that the MAC profiles for organic reference spectra are obtained within an arbitrary scaling factor and, for consistency, we assume the same for ammonium sulfate. 
As we discuss below, the arbitrary scaling constraint is sufficient for our evaluation. The data are illustrated in Appendix \ref{app:IFS}.

Let $\mathbf{y}^{k}$ denote a filter for a target analyte $k$ and denote the amount of the  analyte present on the filter by $f^{k}$. The filter is modeled by 
$\mathbf{y}^{k}\!=\!\mathbf{X}^{k}\v{w}^{k}$, where the first $c\!-\!1$ columns of $\mathbf{X}^{k}\!\in\!\mathbb{R}^{W \times c}$ represent a basis for the interference, the last column contains the spectral profile for analyte $k$ and $W$ is the number of wavenumbers.  
Due to the arbitrary scaling of the reference profile, the coefficient $w^{k}_{c}$  represents a value proportional to the amount of the target analyte in the filter so that $f^{k}\!=\!\beta^{k} w^{k}_{c}$, with $\beta^{k}$ independent of $\mathbf{y}^{k},\v{w}^{k}$. 
To estimate $\beta^{k}$, we first find $\hat{\v{w}}_{i}^{k}$ by applying a chosen estimation method to the model $\mathbf{y}_{i}^{k}\!=\!\mathbf{X}^{k}\boldsymbol{w}_{i}^{k}$ over a batch of filters 
$\{ \mathbf{y}_{i}^{k}: i\!=\!1,2,..,N_{k} \}$. 
We compute  $\hat{\beta}^{k}$ as the coefficient  of a no-intercept univariate regression, with $f_{i}^{k}$ as targets and $\hat{w}_{ic}^{k}$ as inputs. 
The predicted amount of the analyte present in a new filter $\mathbf{y}^{k}$ is $\hat{\beta}^{k}\hat{w}_{c}$, where $\hat{\v{w}}$ is obtained by applying the chosen estimation method to the model $\mathbf{y}^{k}\!=\!\mathbf{X}^{k}\v{w}^{k}$.

We compare two procedures to fit the model $\mathbf{y}^{k}\!=\!\mathbf{X}^{k}\v{w}^{k}$, where the first procedure is based on the linear extended multiplicative signal correction (EMSC) model \citep{Kohler2008}. 
Here the basis for the interference is taken as the leading $c-1$ right singular vectors taken from the singular value decomposition of the matrix obtained by a row-wise stacking of the blank filters, and the coefficients are estimated using ordinary least squares (OLS). 
We treat the number of singular vectors as a hyper-parameter and, since the predictive accuracy of our model is measured by the $R^{2}$-score associated with the univariate regression model, we simply chose the number of singular vectors as the number that maximizes this statistic.

For the second procedure, we consider an approach based on OTLM. 
The interference is modeled as a linear combination of all available blank filters, without orthogonalization, and  Algorithm~\ref{alg:otlm} ($\epsilon\!=\!1.0$, no penalty and $\ell_{2}$ loss for the datafit term) is applied to the model  $\mathbf{y}^{k} = \mathbf{X}^{k}\v{w}^{k}$. 
For the cost matrix, we used $C_{ij}\!=\!0.01\cdot|\omega_{i} -\omega_{j}|$ where $\omega_{i}$ denotes the $i$th wavenumber. 
In addition, we found it beneficial to incorporate information about the reference spectral profile into the cost matrix $\mathbf{C}$, the idea being to be strict to not allow mass around one peak area in the reference profile, in the source marginal, to be transported to other regions in the target marginal, and vice versa.
More formally, for an analyte $k$ with spectral profile $\v{\nu}^{k}$, we specify a threshold $t$, compute $ \mathcal{A} =  \{j: \nu^k_{j} \geq t  \}$ and update $C_{ij}\!=\!C_{ji} = \infty$ for all $i \in \mathcal{A}$ such that $j$ is not connected to $i$ in $\mathcal{A}$.\footnote{For example, $j >i$ is connected to $i$ in $\mathcal{A}$ if $i,i+1,..,j \in \mathcal{A}$.} 
We treat $\lambda,t$ as hyper-parameters set to the values given in Table \ref{tab:IR_OTLM_PARAMS}. 
For a chosen analyte, these were chosen from the grids $\lambda \in \{300,350,400,450,500 \}$ and $t \in \{0.1 \cdot j: j = 1,2,...,9 \}$ by randomly selecting a single blank filter to use as a basis for the intereference, and selecting the hyper-parameter configuration that most frequently maximizes the $R^{2}$-score of the univariate regression when OTLM is run in a leave-one-out (LOO) manner. 

For evaluation, we consider LOO predictions where, for a chosen target analyte, the amount of the analyte present in a filter is predicted by a model trained on the other filters only.
We use the $R^{2}$-score as the evaluation metric. 
In the experiments we find that the EMSC model offers little room for improvement when all the blank filters are used.  
However, when using a limited number of blank filters, OTLM does offer improvement over the EMSC model. 
Hence, the proposed methodology is expected to be useful in applications where only a limited number of relevant examples are available to describe the interference. 
Examples of blank filter spectra are considered most relevant when they are obtained from the same manufactured lot of filters; obtaining many such example spectra is often prohibitive. 
We illustrate this in Table \ref{tab:IR_results} where we report the mean and standard deviation of the LOO $R^{2}$ values obtained by randomly drawing blank filters $B \in \{1,2,3,4,5 \}$, 10 times. 
Our results show OTLM is beneficial to use when less than four blank filters are available.

\begin{table}[]
    \centering
    \begin{tabular}{c c c c c c}
       \hline
Hyper-  & 12- &	Ammonium &	Fruc- &	Levo- &	Suberic  \\
parameter &   Tricosanone             &   sulfate  & tose          &   glucosan   & Acid \\
\hline 
$\lambda$ & 400 & 500 & 500 & 500& 450 \\
$t$     & 0.8 & 0.2 & 0.5 & 0.4& 0.3 \\
\hline 
    \end{tabular}
    \caption{Hyper-parameters selected for OTLM in the prediction of the 
    amount of the different analytes.}
    \label{tab:IR_OTLM_PARAMS}
  \small
    \centering
\resizebox{0.95\textwidth}{!}{
    \begin{tabular}{c  c  c  c  c  c  c }
     \hline
Sample & Method & 12- &	Ammonium &	Fruc- &	Levo- &	Suberic  \\
Size      &        &   Tricosanone             &   sulfate  & tose          &   glucosan            & Acid \\
     \hline
1& EMSC&	97.26 $\pm$ 0.7&	96.4 $\pm$ 0.79&	96.55 $\pm$ 0.68&	63.24 $\pm$ 11.13&	62.12 $\pm$ 3.3 \\
& OTLM&	\textbf{99.29 $\pm$ 0.02}&	\textbf{99.14 $\pm$ 0.04}&	\textbf{98.01 $\pm$ 0.31}&	\textbf{92.78 $\pm$ 2.68}&	\textbf{86.56 $\pm$ 6.59} \\
\hline
2& EMSC&	98.55 $\pm$ 1.06&	97.39 $\pm$ 1.43&	91.04 $\pm$ 7.91&	85.94 $\pm$ 13.73&	71.58 $\pm$ 17.08 \\
& OTLM&	\textbf{99.34 $\pm$ 0.06}&	\textbf{99.15 $\pm$ 0.09}&	\textbf{97.99 $\pm$ 0.39}&	\textbf{90.6 $\pm$ 2.66}&	\textbf{88.41 $\pm$ 7.39} \\
\hline
3& EMSC&	98.75 $\pm$ 1.01&	98.83 $\pm$ 0.97&	96.65 $\pm$ 1.65&	\textbf{93.72 $\pm$ 3.98}&	80.04 $\pm$ 17.12 \\
& OTLM&	\textbf{99.34 $\pm$ 0.04}&	\textbf{99.25 $\pm$ 0.08}&	\textbf{98.15 $\pm$ 0.17}&	90.68 $\pm$ 2.35&	\textbf{91.37 $\pm$ 2.48} \\
\hline
4& EMSC&	\textbf{99.59 $\pm$ 0.43}&	\textbf{99.31 $\pm$ 0.24}&	88.69 $\pm$ 16.19&	87.17 $\pm$ 16.83&	\textbf{96.13 $\pm$ 3.76} \\
& OTLM&	99.35 $\pm$ 0.05&	99.24 $\pm$ 0.1&	\textbf{98.17 $\pm$ 0.18}&	\textbf{91.32} $\pm$ 2.34&	91.73 $\pm$ 1.34 \\
\hline
5& EMSC&	\textbf{99.80 $\pm$ 0.01}&	\textbf{99.44 $\pm$ 0.12}&	97.11 $\pm$ 2.19&	\textbf{91.84 $\pm$ 5.31}&	\textbf{98.75 $\pm$ 0.92} \\
& OTLM&	99.37 $\pm$ 0.05 &	99.28 $\pm$ 0.09 & \textbf{98.08 $\pm$ 0.16}  &	91.56 $\pm$ 2.38 &	90.37 $\pm$ 3.7	 \\
    \hline
    \end{tabular} }
    \caption{Mean and standard deviation of the $R^{2}$ values (expressed as percentages) of the leave-one-out predictions over the random sub-samples taken from the blank filters. The results suggest that using OTLM over EMSC is beneficial up to 3 blank filters. }
    \label{tab:IR_results}
\end{table}

\section{Conclusions}

We present a scaling algorithm for solving optimal transport linear regression models with various convex datafit and penalty terms, under the entropic regularization.
The iterative algorithm consists of simple multiplicative Sinkhorn-like updates.
It offers good prospects for regression problems where the noise can be understood as a perturbation of ``shape'', rather than additive or multiplicative per-sample noise.
The simplicity and scalability of this algorithm opens possibilities for more detailed studies of optimal transport linear models, in terms of their convergence and asymptotic properties for different problems.
We demonstrate the applicability of this method for analyses of spectroscopic data, and we expect it to be useful for a range of additional real-world applications.

\section*{Acknowledgments}
We would like to thank Ilnura Usmanova and William Aeberhard for very useful comments on the manuscript.
The work has been supported by the Swiss Data Science Center collaborative grant ``Smart Analysis of MUonic x-Rays with Artificial Intelligence'' and ``Cost-effective chemical speciation monitoring of particulate matter air pollution''.
The MIXE data was created within the Sinergia project ``Deep$\mu$'' funded by the Swiss National Science Foundation, grant number: 193691.
The data for infrared spectroscopy was collected as part of the IMPROVE program with support from the US Environmental Protection Agency (National Park Service cooperative agreement Pl8AC01222).

\bibliography{otlm}

\begin{thebibliography}{42}
\providecommand{\natexlab}[1]{#1}
\providecommand{\url}[1]{\texttt{#1}}
\expandafter\ifx\csname urlstyle\endcsname\relax
  \providecommand{\doi}[1]{doi: #1}\else
  \providecommand{\doi}{doi: \begingroup \urlstyle{rm}\Url}\fi

\bibitem[Arjovsky et~al.(2017)Arjovsky, Chintala, and
  Bottou]{arjovsky2017:wgan}
Martin Arjovsky, Soumith Chintala, and L{\'e}on Bottou.
\newblock {W}asserstein generative adversarial networks.
\newblock In \emph{Proceedings of the 34th International Conference on Machine
  Learning}, volume~70 of \emph{Proceedings of Machine Learning Research},
  pages 214--223. PMLR, 06--11 Aug 2017.

\bibitem[Benamou(2003)]{benamou2003:numerical}
Jean-David Benamou.
\newblock Numerical resolution of an “unbalanced” mass transport problem.
\newblock \emph{ESAIM: Mathematical Modelling and Numerical Analysis},
  37\penalty0 (5):\penalty0 851--868, 2003.

\bibitem[Benamou et~al.(2015)Benamou, Carlier, Cuturi, Nenna, and
  Peyr{\'e}]{benamou2015:iterative}
Jean-David Benamou, Guillaume Carlier, Marco Cuturi, Luca Nenna, and Gabriel
  Peyr{\'e}.
\newblock Iterative bregman projections for regularized transportation
  problems.
\newblock \emph{SIAM Journal on Scientific Computing}, 37\penalty0
  (2):\penalty0 A1111--A1138, 2015.

\bibitem[Bernton et~al.(2022)Bernton, Ghosal, and Nutz]{bernton2022:entropic}
Espen Bernton, Promit Ghosal, and Marcel Nutz.
\newblock Entropic optimal transport: Geometry and large deviations.
\newblock \emph{Duke Mathematical Journal}, 171\penalty0 (16):\penalty0
  3363--3400, 2022.

\bibitem[Bertrand et~al.(2022)Bertrand, Klopfenstein, Bannier, Gidel, and
  Massias]{bertrand2022:skglm}
Quentin Bertrand, Quentin Klopfenstein, Pierre-Antoine Bannier, Gauthier Gidel,
  and Mathurin Massias.
\newblock Beyond l1: Faster and better sparse models with skglm.
\newblock \emph{Advances in Neural Information Processing Systems},
  35:\penalty0 38950--38965, 2022.

\bibitem[Biswas et~al.(2023)Biswas, Megatli-Niebel, Raselli, Simke, Cocolios,
  Deokar, Elender, Gerchow, Hess, Khasanov, Knecht, Luetkens, Ninomiya, Papa,
  Prokscha, Reiter, Sato, Severijns, Shiroka, Seidlitz, Vogiatzi, Wang,
  Wauters, Warr, and Amato]{Biswas2023:fibula}
Sayani Biswas, Isabel Megatli-Niebel, Lilian Raselli, Ronald Simke,
  Thomas~Elias Cocolios, Nilesh Deokar, Matthias Elender, Lars Gerchow, Herbert
  Hess, Rustem Khasanov, Andreas Knecht, Hubertus Luetkens, Kazuhiko Ninomiya,
  Angela Papa, Thomas Prokscha, Peter Reiter, Akira Sato, Nathal Severijns,
  Toni Shiroka, Michael Seidlitz, Stergiani~Marina Vogiatzi, Chennan Wang,
  Frederik Wauters, Nigel Warr, and Alex Amato.
\newblock The non-destructive investigation of a late antique knob bow fibula
  (b{\"u}gelknopffibel) from kaiseraugst/ch using muon induced x-ray emission
  (mixe).
\newblock \emph{Heritage Science}, 11\penalty0 (1):\penalty0 43, 2023.

\bibitem[Chapel et~al.(2021)Chapel, Flamary, Wu, F{\'e}votte, and
  Gasso]{chapel2021:unbalanced}
Laetitia Chapel, R{\'e}mi Flamary, Haoran Wu, C{\'e}dric F{\'e}votte, and
  Gilles Gasso.
\newblock Unbalanced optimal transport through non-negative penalized linear
  regression.
\newblock \emph{Advances in Neural Information Processing Systems},
  34:\penalty0 23270--23282, 2021.

\bibitem[Chizat et~al.(2018)Chizat, Peyr{\'e}, Schmitzer, and
  Vialard]{chizat2018:scaling}
Lenaic Chizat, Gabriel Peyr{\'e}, Bernhard Schmitzer, and Fran{\c{c}}ois-Xavier
  Vialard.
\newblock Scaling algorithms for unbalanced optimal transport problems.
\newblock \emph{Mathematics of Computation}, 87\penalty0 (314):\penalty0
  2563--2609, 2018.

\bibitem[Ciach et~al.(2024)Ciach, Miasojedow, Skoraczy{\'n}ski, Majewski,
  Startek, Valkenborg, and Gambin]{ciach2024:masserstein}
Micha{\l}~Aleksander Ciach, B{\l}a{\.z}ej Miasojedow, Grzegorz
  Skoraczy{\'n}ski, Szymon Majewski, Micha{\l} Startek, Dirk Valkenborg, and
  Anna Gambin.
\newblock Masserstein: Linear regression of mass spectra by optimal transport.
\newblock \emph{Rapid Communications in Mass Spectrometry}, \penalty0
  (n/a):\penalty0 e8956, 2024.
\newblock \doi{https://doi.org/10.1002/rcm.8956}.
\newblock e8956 rcm.8956.

\bibitem[Cominetti and Mart{\'\i}n(1994)]{cominetti1994:asymptotic}
Roberto Cominetti and J~San Mart{\'\i}n.
\newblock Asymptotic analysis of the exponential penalty trajectory in linear
  programming.
\newblock \emph{Mathematical Programming}, 67:\penalty0 169--187, 1994.

\bibitem[Cuturi(2013)]{cuturi2013:sinkhorn}
Marco Cuturi.
\newblock Sinkhorn distances: Lightspeed computation of optimal transport.
\newblock \emph{Advances in neural information processing systems}, 26, 2013.

\bibitem[Earle et~al.(2006)Earle, Pancescu, Cosic, Zasetsky, and
  Sloan]{Earle2006}
M.~E. Earle, R.~G. Pancescu, B.~Cosic, A.~Y. Zasetsky, and J.~J. Sloan.
\newblock Temperature-{Dependent} {Complex} {Indices} of {Refraction} for
  {Crystalline} ({NH}4)2so4.
\newblock \emph{The Journal of Physical Chemistry A}, 110\penalty0
  (48):\penalty0 13022--13028, December 2006.
\newblock ISSN 1089-5639.
\newblock \doi{10.1021/jp064704s}.
\newblock URL \url{https://doi.org/10.1021/jp064704s}.

\bibitem[Erb(2024)]{erb2024:pybaselines}
Donald Erb.
\newblock {pybaselines}: A {Python} library of algorithms for the baseline
  correction of experimental data, 2024.
\newblock URL \url{https://github.com/derb12/pybaselines}.

\bibitem[Flamary et~al.(2016)Flamary, F{\'e}votte, Courty, and
  Emiya]{flamary2016:music}
R{\'e}mi Flamary, C{\'e}dric F{\'e}votte, Nicolas Courty, and Valentin Emiya.
\newblock Optimal spectral transportation with application to music
  transcription.
\newblock \emph{Advances in Neural Information Processing Systems}, 29, 2016.

\bibitem[Foreman-Mackey et~al.(2013)Foreman-Mackey, Hogg, Lang, and
  Goodman]{foreman2013emcee}
Daniel Foreman-Mackey, David~W Hogg, Dustin Lang, and Jonathan Goodman.
\newblock emcee: the mcmc hammer.
\newblock \emph{Publications of the Astronomical Society of the Pacific},
  125\penalty0 (925):\penalty0 306, 2013.

\bibitem[Frogner et~al.(2015)Frogner, Zhang, Mobahi, Araya, and
  Poggio]{frogner2015:domainadaptation}
Charlie Frogner, Chiyuan Zhang, Hossein Mobahi, Mauricio Araya, and Tomaso~A
  Poggio.
\newblock Learning with a wasserstein loss.
\newblock \emph{Advances in neural information processing systems}, 28\penalty0
  (9):\penalty0 1853--1865, 2015.
\newblock \doi{10.1109/TPAMI.2016.2615921}.

\bibitem[{GEEL}(2015)]{geel:bcr961}
{European Commission Joint Research Centre Institute for Reference Materials
  and Measurements} {GEEL}.
\newblock {BCR-961} copper alloy standard, 2015.
\newblock URL
  \url{https://crm.jrc.ec.europa.eu/p/q/BCR-69/BCR-691-COPPER-ALLOYS/BCR-691}.

\bibitem[Gerchow et~al.(2023)Gerchow, Biswas, Janka, Vigo, Knecht, Vogiatzi,
  Ritjoho, Prokscha, Luetkens, and Amato]{gerchow:2022mixe}
Lars Gerchow, Sayani Biswas, Gianluca Janka, Carlos Vigo, Andreas Knecht,
  Stergiani~Marina Vogiatzi, Narongrit Ritjoho, Thomas Prokscha, Hubertus
  Luetkens, and Alex Amato.
\newblock {GermanIum array for non-destructive testing (GIANT) setup for
  muon-induced x-ray emission (MIXE) at the Paul Scherrer Institute}.
\newblock \emph{Rev. Sci. Instrum.}, 94\penalty0 (4):\penalty0 045106, 2023.
\newblock \doi{10.1063/5.0136178}.

\bibitem[Goulart and Chen(2024)]{goulart2024:clarabel}
Paul~J Goulart and Yuwen Chen.
\newblock Clarabel: An interior-point solver for conic programs with quadratic
  objectives.
\newblock \emph{arXiv preprint arXiv:2405.12762}, 2024.

\bibitem[Hahn et~al.(2024)Hahn, Zsombor-Pindera, Kennepohl, and
  DeBeer]{hahn2024spectrafit}
Anselm~W Hahn, Joseph Zsombor-Pindera, Pierre Kennepohl, and Serena DeBeer.
\newblock Introducing spectrafit: An open-source tool for interactive spectral
  analysis.
\newblock \emph{ACS omega}, 9\penalty0 (22):\penalty0 23252--23265, 2024.

\bibitem[Hofmann et~al.(2023)Hofmann, Schreyer, Biswas, Gerchow, Wiebe,
  Schumann, Lindemann, Garc{\'\i}a, Lanari, Gfeller, Vigo, Das, Hotz, {von
  Schoeler}, Ninomiya, Niikura, Ritjoho, and Amato]{hofmann2023:arrowhead}
Beda~A. Hofmann, Sabine~Bolliger Schreyer, Sayani Biswas, Lars Gerchow, Daniel
  Wiebe, Marc Schumann, Sebastian Lindemann, Diego~Ram{\'\i}rez Garc{\'\i}a,
  Pierre Lanari, Frank Gfeller, Carlos Vigo, Debarchan Das, Fabian Hotz,
  Katharina {von Schoeler}, Kazuhiko Ninomiya, Megumi Niikura, Narongrit
  Ritjoho, and Alex Amato.
\newblock An arrowhead made of meteoritic iron from the late bronze age
  settlement of m{\"o}rigen, switzerland and its possible source.
\newblock \emph{Journal of Archaeological Science}, 157:\penalty0 105827, 2023.

\bibitem[Huangfu and Hall(2018)]{huangfu2018:highs}
Qi~Huangfu and JA~Julian Hall.
\newblock Parallelizing the dual revised simplex method.
\newblock \emph{Mathematical Programming Computation}, 10\penalty0
  (1):\penalty0 119--142, 2018.

\bibitem[{John Wiley \& Sons, Inc.}(2025)]{SpectraBase}
{John Wiley \& Sons, Inc.}
\newblock Spectrabase.
\newblock \url{https://spectrabase.com/}, 2025.
\newblock Accessed: 2025-03-02.

\bibitem[Johnson et~al.(2009)Johnson, Bickson, and Dolev]{johnson2009fixing}
Jason~K Johnson, Danny Bickson, and Danny Dolev.
\newblock Fixing convergence of {G}aussian belief propagation.
\newblock In \emph{2009 IEEE International Symposium on Information Theory},
  pages 1674--1678. IEEE, 2009.

\bibitem[Kantorovich(1960)]{kantorovich1960:wasserstein}
Leonid~V Kantorovich.
\newblock Mathematical methods of organizing and planning production.
\newblock \emph{Management science}, 6\penalty0 (4):\penalty0 366--422, 1960.

\bibitem[Kohler et~al.(2008)Kohler, Sulé-Suso, Sockalingum, Tobin, Bahrami,
  Yang, Pijanka, Dumas, Cotte, van Pittius, Parkes, and Martens]{Kohler2008}
A.~Kohler, J.~Sulé-Suso, G.~D. Sockalingum, M.~Tobin, F.~Bahrami, Y.~Yang,
  J.~Pijanka, P.~Dumas, M.~Cotte, D.~G. van Pittius, G.~Parkes, and H.~Martens.
\newblock Estimating and {Correcting} {Mie} {Scattering} in
  {Synchrotron}-{Based} {Microscopic} {Fourier} {Transform} {Infrared}
  {Spectra} by {Extended} {Multiplicative} {Signal} {Correction}.
\newblock \emph{Applied Spectroscopy}, 62\penalty0 (3):\penalty0 259--266,
  March 2008.
\newblock ISSN 0003-7028.
\newblock \doi{10.1366/000370208783759669}.
\newblock URL \url{https://doi.org/10.1366/000370208783759669}.

\bibitem[Lee and Seung(2000)]{lee2000:nnmf}
Daniel Lee and H~Sebastian Seung.
\newblock Algorithms for non-negative matrix factorization.
\newblock \emph{Advances in neural information processing systems}, 13, 2000.

\bibitem[Montesuma et~al.(2024)Montesuma, Mboula, and
  Souloumiac]{montesuma2024:otml}
Eduardo~Fernandes Montesuma, Fred Maurice~Ngol{\`e} Mboula, and Antoine
  Souloumiac.
\newblock Recent advances in optimal transport for machine learning.
\newblock \emph{IEEE Transactions on Pattern Analysis and Machine
  Intelligence}, 2024.

\bibitem[Nakhostin et~al.(2016)Nakhostin, Courty, Flamary, and
  Corpetti]{nakhostin2016:supervised}
Sina Nakhostin, Nicolas Courty, R{\'e}mi Flamary, and Thomas Corpetti.
\newblock Supervised planetary unmixing with optimal transport.
\newblock In \emph{2016 8th Workshop on Hyperspectral Image and Signal
  Processing: Evolution in Remote Sensing (WHISPERS)}, pages 1--5. IEEE, 2016.

\bibitem[Nelder and Wedderburn(1972)]{nelder1972glm}
John~Ashworth Nelder and Robert~WM Wedderburn.
\newblock Generalized linear models.
\newblock \emph{Journal of the Royal Statistical Society Series A: Statistics
  in Society}, 135\penalty0 (3):\penalty0 370--384, 1972.

\bibitem[Newville et~al.(2016)Newville, Stensitzki, Allen, Rawlik, Ingargiola,
  and Nelson]{newville2016lmfit}
Matthew Newville, Till Stensitzki, Daniel~B Allen, Michal Rawlik, Antonino
  Ingargiola, and Andrew Nelson.
\newblock Lmfit: Non-linear least-square minimization and curve-fitting for
  python.
\newblock \emph{Astrophysics Source Code Library}, pages ascl--1606, 2016.

\bibitem[Onken et~al.(2021)Onken, Fung, Li, and Ruthotto]{onken2021otflow}
Derek Onken, Samy~Wu Fung, Xingjian Li, and Lars Ruthotto.
\newblock Ot-flow: Fast and accurate continuous normalizing flows via optimal
  transport.
\newblock In \emph{Proceedings of the AAAI Conference on Artificial
  Intelligence}, volume~35, pages 9223--9232, 2021.

\bibitem[Peyr{\'e} et~al.(2019)Peyr{\'e}, Cuturi,
  et~al.]{peyre2019:computational}
Gabriel Peyr{\'e}, Marco Cuturi, et~al.
\newblock Computational optimal transport: With applications to data science.
\newblock \emph{Foundations and Trends{\textregistered} in Machine Learning},
  11\penalty0 (5-6):\penalty0 355--607, 2019.

\bibitem[Qu{\'e}rel et~al.(2025)Qu{\'e}rel, Biswas, Heiss, Gerchow, Wang,
  Asakura, M{\"u}ller, Das, Guguchia, Hotz, et~al.]{querel2025:battery}
Edouard Qu{\'e}rel, Sayani Biswas, Michael~W Heiss, Lars Gerchow, Qing Wang,
  Ryo Asakura, Gian M{\"u}ller, Debarchan Das, Zurab Guguchia, Fabian Hotz,
  et~al.
\newblock Overcoming the probing-depth dilemma in spectroscopic analyses of
  batteries with muon-induced x-ray emission (mixe).
\newblock \emph{Journal of Materials Chemistry A}, 2025.

\bibitem[Ruthenburg et~al.(2014)Ruthenburg, Perlin, Liu, McDade, and
  Dillner]{Ruthenburg2014}
Travis~C. Ruthenburg, Pesach~C. Perlin, Victor Liu, Charles~E. McDade, and
  Ann~M. Dillner.
\newblock Determination of organic matter and organic matter to organic carbon
  ratios by infrared spectroscopy with application to selected sites in the
  improve network.
\newblock \emph{Atmospheric Environment}, 86:\penalty0 47--57, April 2014.
\newblock \doi{10.1016/j.atmosenv.2013.12.034}.

\bibitem[S{\'e}journ{\'e} et~al.(2023)S{\'e}journ{\'e}, Peyr{\'e}, and
  Vialard]{sejourne2023:unbalanced}
Thibault S{\'e}journ{\'e}, Gabriel Peyr{\'e}, and Fran{\c{c}}ois-Xavier
  Vialard.
\newblock Unbalanced optimal transport, from theory to numerics.
\newblock \emph{Handbook of Numerical Analysis}, 24:\penalty0 407--471, 2023.

\bibitem[Sinkhorn(1964)]{sinkhorn1964:relationship}
Richard Sinkhorn.
\newblock A relationship between arbitrary positive matrices and doubly
  stochastic matrices.
\newblock \emph{Annals of Mathematical Statistics}, 35:\penalty0 876--879,
  1964.

\bibitem[Sturniolo and Hillier(2021)]{sturniolo2021:mudirac}
Simone Sturniolo and Adrian Hillier.
\newblock Mudirac: A dirac equation solver for elemental analysis with muonic
  x-rays.
\newblock \emph{X-Ray Spectrometry}, 50\penalty0 (3):\penalty0 180--196, 2021.

\bibitem[Tibshirani(1996)]{tibshirani1996:regression}
Robert Tibshirani.
\newblock Regression shrinkage and selection via the lasso.
\newblock \emph{Journal of the Royal Statistical Society Series B: Statistical
  Methodology}, 58\penalty0 (1):\penalty0 267--288, 1996.

\bibitem[Villani(2009)]{villani2009:optimal}
C{\'e}dric Villani.
\newblock \emph{Optimal transport: old and new}, volume 338.
\newblock Springer, 2009.

\bibitem[Weed(2018)]{weed2018:explicit}
Jonathan Weed.
\newblock An explicit analysis of the entropic penalty in linear programming.
\newblock In \emph{Conference On Learning Theory}, pages 1841--1855. PMLR,
  2018.

\bibitem[Zou and Hastie(2005)]{zou2005:elastic}
Hui Zou and Trevor Hastie.
\newblock Regularization and variable selection via the elastic net.
\newblock \emph{Journal of the Royal Statistical Society Series B: Statistical
  Methodology}, 67\penalty0 (2):\penalty0 301--320, 2005.

\end{thebibliography}

\newpage
\appendix

\section{Appendix: Proximal Operators for Non-Negative Span Projection}
\label{app:prox_nnsx} 
The gradient of the majorized objective in Equation~\ref{eqn:obj_nnsx_majorized} is given by 
\begin{equation*}
\frac{\partial G}{\partial w_j}(\v{w, w'})
=
\sum_{i=1}^N
X_{ij} \,\ln\left(\frac{X_{ij} w_j}{Z_{ij}}\frac{1}{y_i}\right) + \frac{\partial R(w_j)}{\partial w_j}.
\end{equation*}
Setting the gradient to zero and rearranging gives:
\begin{equation*}
\sum_{i=1}^N X_{ij}  \ln \frac{X_{ij} y_i}{Z_{ij} w_j}
=
- \frac{\partial R(w_j)}{\partial w_j}
\end{equation*}
The term $\frac{\partial R(w_j)}{\partial w_j}$ is [0, $\alpha$, $\alpha w_j$, $\alpha + \beta w_j$] for the [unpenalized, LASSO, Ridge, Elastic Net] penalties.
The solution is straightforward in the case of the $\ell_1$ penalty.
For the squared $\ell_2$ penalty, we define:
\begin{equation*}
x_j 
\;=\; 
\sum_{i=1}^N X_{ij}, 
\quad
d_j
\;=\; 
\sum_{i=1}^N
X_{ij}
\bigl[
  \ln(X_{ij}) 
  - \ln\bigl(Z_{ij}\bigr)
  - \ln\bigl(y_i\bigr)
\bigl].
\end{equation*}
which gives:
\begin{equation*}
w_j
\;=\;
\exp\Bigl(-\tfrac{d_j}{x_j}\Bigr)
\;\exp\Bigl(-\tfrac{\alpha}{\epsilon}\tfrac{w_j}{x_j}\Bigr).
\end{equation*}
After some manipulation, this equation can be expressed as a Lambert W form, leading to the result in Table~\ref{tab:prox_nnsx}.
For the Elastic Net penalty, we have:
\begin{equation*}
  \epsilon \sum_{i=1}^N X_{ij}\,\ln\!\Bigl(\tfrac{X_{ij}\,w_j}{Z_{ij}\,y_i}\Bigr)
  \;+\;
  \alpha
  \;+\;
  \beta\,w_j
  \;=\;
  0.
\end{equation*}
Rearranging and isolating $\ln(w_j)$ leads to an equation of the form $\ln(w_j) + \gamma_j \, w_j = \nu_j$, with 
\begin{equation*}
\gamma_j
=
\frac{\beta}{\epsilon} \frac{1}{x_j},
\quad
\nu_j
=
-\frac{1}{x_j}
\left(
  \frac{\alpha}{\epsilon} 
  + \sum_{i=1}^N X_{ij}\,\ln\!\left(\tfrac{X_{ij}}{Z_{ij}\,y_i}\right)
\right),
\end{equation*}
which leads to the Lambert W function solution in Table~\ref{tab:prox_nnsx}.

\section{Appendix: Derivation of Proximal Operators for Additional Marginal Datafits}
\label{app:l2marginal}

To calculate the updates for various marignal datafit functions, we need to find the proximal operator for the marginal datafit functions with respect to the KL divergence, see Equation~\ref{eqn:prox}.
A numerical solution for the squared error marginal loss was previously derived by \cite{benamou2003:numerical}, but for a gradient-based approach to UOT.
Due to the multiplicative nature of the scaling algorithm and the non-negativity of the data, we assume that the constraint $\v{Q} \geq 0$ is always satisfied.
Thus, assuming interior solutions and defining $r_j\!= \!\sum_{i=1}^N Q_{ij} \!-\! y_j$, the stationarity condition is:
\begin{equation*}
\begin{aligned} 
\epsilon \log\frac{Q_{ij}}{K_{ij}} + \lambda r_j = 0.
\end{aligned}
\end{equation*}
With $k_j = \sum_{i=1}^N K_{ij}$ and $z_j = \frac{\lambda}{\epsilon} (r_j + y_j)$ we have:
\begin{equation*}
\begin{aligned} 
  z_j e^{z_j} = \frac{\lambda}{\epsilon} k_j \exp\left(\frac{\lambda}{\epsilon} y_j\right).
\end{aligned}
\end{equation*}
By definition of the principal branch of the Lambert W function $\mathrm{W}_0(\cdot)$, we have:
\begin{equation*}
\begin{aligned} 
  z_j = \mathrm{W}_0\left(\frac{\lambda}{\epsilon} k_j \exp \left( \frac{\lambda}{\epsilon}y_j \right)\right).
\end{aligned}
\end{equation*}
Here the Lambert W function is applied element-wise to the vector $z_j$.
This leads to the result in Table~\ref{tab:prox}.

The solution for the Poisson loss is following a similar approach.
Note that we do not use a link function, as in other types of Poisson regression.
The negative log Poisson loss is given by:
$\mathrm{L}(\v{s}|\v{y}) = - \lambda (\v{y}\T \log \v{s} - \v{s})$.
With $q_i \coloneqq \sum_{j=1}^N Q_{ij}$, the stationarity condition for the proximal operator is:
\begin{equation*}
\begin{aligned} 
  \epsilon \log\frac{Q_{ij}}{K_{ij}} - \lambda \frac{y_i}{q_i} - \lambda = 0.
\end{aligned}
\end{equation*}
Let's define $k_{i}\coloneqq\sum_{j=1}^N K_{ij}$ and $u_i \coloneqq \lambda/\epsilon \cdot y_i/q_i$.
This leads to:
\begin{equation*}
\begin{aligned} 
  u_i e^{u_i} =  \frac{\lambda}{\epsilon} \frac{y_i}{k_{i}} \exp\left(\frac{\lambda}{\epsilon}\right).
\end{aligned}
\end{equation*}
Using the Lambert W again, we have:
\begin{equation*}
\begin{aligned} 
  u_i = \mathrm{W}_0\left(\frac{\lambda}{\epsilon} \frac{y_i}{k_{i}}\exp\left(\frac{\lambda}{\epsilon}\right)\right).
\end{aligned}
\end{equation*}
After straightforward manipulations, we can obtain the result in Table~\ref{tab:prox}.

\section{Appendix: Details of the Demonstration Problem}
\label{app:demo}

The transport cost is set to $C_{ij} = \rho |x_i - x_j|$, with $\rho\!=\!0.01$.
The OTLM parameters are $\lambda\!=\!1$, $\epsilon\!=\!0.001$.
For the penalized experiments, the parameters are: $\lambda{=}1$, $\alpha{=}\epsilon{=}0.001$ (Ridge), $\lambda{=}1$, $\alpha{=}0.007$, $\epsilon{=}0.0002$ (LASSO), and $\lambda{=}100$, $\alpha{=}0.0001$, $\epsilon{=}1$ (Poisson).
In all cases, the transport cost is set to $C_{ij} = \rho | x_i - x_j |$, with $\rho = 0.01$.

In the unpenalized case, the TV regression finds weights $\v{w}\!=\!\arg \min_{\v{w}\!\geq\!0} |\v{Xw}\!-\!\v{y}|_1$, while its \ref{eq:otlm} version uses $\mathrm{L}\!=\!\lambda \mathrm{TV}(\cdot|\v{y})$.
The non-negative Ridge problem is defined as $\v{w}\!=\! \arg \min_{\v{w}\!\geq\!0} \|\v{Xw}\!-\!\v{y}\|_2^2\!+\!\alpha \|\v{w}\|_2$, and non-negative LASSO as $\v{w}\!=\! \arg \min_{\v{w}\!\geq\!0} \|\v{Xw}\!-\!\v{y}\|_2^2\!+\!\alpha \|\v{w}\|_1$. The Poisson problem is defined as $\v{w}\!=\! \arg \min_{\v{w}\!\geq\!0} \ - \sum_i^N (\v{y} \odot \log \v{Xw}\!-\!\v{Xw})_i$.
The solution to the TV regression problem is obtained using linear programming solver HiGHS \citep{huangfu2018:highs} and is exact.
The solutions for the Ridge and LASSO problems were obtained using the skglm package \citep{bertrand2022:skglm}. 
The solution for the Poisson problem was obtained using the BFGS solver in scipy.
These solvers were run for very high precision requirements and are very close to exact.

\section{Appendix: Performance Scalability Tests}
\label{app:scaling} 

The performance scalability test is performed on a problem with $N$ samples and $M=N/10$ basis vectors.
The test data is generated using $M$ skew-Gaussian functions $f(x, \mu, \sigma, \gamma)$:
\begin{equation*}
y_i = \sum_{m=1}^M a_m f(x_i, \mu_m, \sigma_m, \gamma_m) + |n_i|,
\end{equation*}
where the 
means $\mu_m$ were randomly sampled with uniform distribution $\mu_m \!\sim\! \mathcal{U}[0,N]$, 
the standard deviations $\sigma_m$ set to  absolute values of samples from normal distribution $\sigma_m \!\sim\! |\mathcal{N}[2, 0.2]|$, 
amplitudes $a_m \!\sim\! |\mathcal{N}[1, 0.2)|+0.01$, 
and skewness $\gamma_m \!\sim\! \mathcal{N}[0, 2]$.
Non-negative random noise was then added, drawn from $n_i \!\sim\! \mathcal{N}[0, 0.002]$.
The basis vectors were created using Gaussian functions with $M$ means at regular intervals from 0 to $N$, $\mu^b_m{=}m \cdot N/M$, standard deviations set to $\sigma^b_m \!=\! 2$, and amplitudes set to $a^b_m\! =\! 1$.
The basis functions were truncated at $\pm 5\sigma^b_m$.

The OTLM problems use:
(i) $\mathrm{L}(\cdot, \cdot) = \frac{1}{2} \ell_2(\cdot\!-\!\cdot)^2$ and $\mathrm{R}(\cdot) = \frac{1}{2} \ell_2(\cdot)^2$ for the OT Ridge,
(ii) $\mathrm{L}(\cdot, \cdot) = \mathrm{TV}(\cdot, \cdot)$ and $\mathrm{R}(\cdot) = \ell_1(\cdot)$ for the OT TV.
The cost function is defined as 
\begin{equation*}
C_{ij} = \begin{cases} \rho |x_i - x_j| & \text{if } |x_i - x_j| < \delta_x^{\mathrm{max}} \\ \infty & \text{otherwise} \end{cases}
\end{equation*}
with $\rho\!=\!0.01$ and $\delta_x^{\mathrm{max}}\!=\!5\sigma^b_m$.
The OTLM parameters used are $\lambda\!=\!1$, $\alpha\!=\!\epsilon\!=\!0.001$.
The cost and basis matrices is kept in a sparse format.
The linear OTLM without entropic regularization is solved using the scipy.optimize.linprog rountine, running the HiGHS solver \citep{huangfu2018:highs} at default settings.
The L2 OTLM without entropic regularization is solved using the Clarabel solver \citep{goulart2024:clarabel}, via the CVXPY interface, with default settings.
Both methods utilized sparse matrix representations.

\section{Muonic X-ray Spectra}
\label{app:muonic_xray_spec}

The muonic X-ray analysis has recently emerged as a powerful tool for elemental analysis, with applications to operando experiments on batteries \citep{querel2025:battery}, studies of archeological artifacts \citep{Biswas2023:fibula}, extraterrestrial materials \citep{hofmann2023:arrowhead}, and others.

The samples were chosen to be metal alloy certified reference materials: Copper alloys BCR-961 A,B,C,D,E\footnote{\url{https://crm.jrc.ec.europa.eu/p/q/BCR-69/BCR-691-COPPER-ALLOYS/BCR-691}}, 
The data used in this work was collected with the muon-induced X-ray emission (MIXE) instrument at the Paul Scherrer Institute \citep{gerchow:2022mixe} in the 2023-2024 measurement campaigns.
The energies of recorded X-ray events were calibrated using radioactive sources \citep{gerchow:2022mixe}.
The events were converted into a histogram from 70-6100 keV in in narrow energy bins of 0.305 keV from all detector channels.  
Then, the spectra baselines were modelled using the \textsc{PyBaselines} package \citep{erb2024:pybaselines}, using the ``Iterative Reweighted Spline Quantile Regression'' (IRSQR) algorithm. 
The energy efficiency of the detectors can be modeled in a reasonable approximation by the baseline, due to the fairly flat energy distribution of Compton scattering events
Therefore, to obtain the energy-efficiency corrected spectra, we divide the histogram by the baseline. 

We will describe the case for the BCR-691-A sample; other samples were analyzed the same way.
Considering elements $Z \in { \{ \mathrm{Cu, Zn, Sn, Pb} \} }$, the dictionary $\v{X}$ contains two basis vectors for each element, corresponding to the $K_{\alpha 1}$ and $K_{\alpha 2}$ lines.
Each of them is created as a sum of Gaussian functions corresponding to all naturally abundant isotopes of the element.
The X-ray energy levels $e^{K_{\alpha 1}}_{ZA}, e^{K_{\alpha 2}}_{ZA}$ for $K_{\alpha 1}, K_{\alpha 2}$ transitions calculated by the \textsc{Mudirac} code \citep{sturniolo2021:mudirac} for each element $Z$ and isotope $A$.
The energy spread of the lines are taken from calibration measurements of MIXE \citep[Figure 9c]{gerchow:2022mixe}, which calculated the standard deviation function of the line as a function of its energy $e$ in keV $\sigma: \mathbb{R}_{+} \rightarrow \mathbb{R}_{+}$. 
The measured points are fitted with a linear function resulting in $\sigma (e)\!=\!0.0003\!+\!0.4983 e$
The Gaussian basis functions for element $Z$ and $K_{\alpha t}, \ t\!=\!1,2$ are supported on $N\!=\!8615$ energy levels of the grid $\v{e} \in \mathbb{R}^N$:
\begin{equation*}
  X_{Zt} = \sum_{A \in \mathrm{isotopes}(Z)} a_{Z,A} \cdot \mathcal{N}( \v{e} \ | \ e_{Z,A}^{K_{\alpha t}}, \sigma(e_{Z,A}^{K_{\alpha t}}))
\end{equation*}
where $\mathcal{N}(\cdot | \cdot, \cdot)$ is the Gaussian distribution. 
Each basis column is normalized to have unit mass.
The columns are stacked horizontally to form the dictionary $\v{X} \in \mathbb{R}^{N \times 36}$.
Note that this relation could be improved by performing a dedicated peak spread measurement for each sample; this would require an additional multi-step procedure. 
As the goal of this work is to demonstrate the robustness of OTLM to such inaccuracies, we do not pursue this here.
We use the TV loss and a small $\ell_1$ penalty.
The OT cost is defined as 
$$C_{ij} = \begin{cases}
  \rho (x_i\!-\!x_j)^2 / x_i^2 & \text{if } |x_i\!-\!x_j| \leq \Delta x^{\mathrm{max}} \\
  \infty & \text{if } \text{otherwise}
\end{cases}$$
with $\Delta x^{\mathrm{max}}\!=\!10$ keV, $\rho\!=\!5\cdot10^5$, $\alpha\!=\!\epsilon\!=\!10^{-1}$, $\lambda\!=\!1$.
The non-stationary cost is needed, as the energy measurement error is increasing with the energy itself.
The entropic regularization $\epsilon$ value is chosen to be as small as possible, and the transport cost as large as possible, such that the results are numerically stable.
The parameter $\lambda$ is set to be as small as possible, while allowing transport to happen: we required that the marginal fits the target very well.
Overall, we find that the results were not sensitive to these choices as long as the $\lambda$ was large enough to allow fitting the target marginals well.
The final weights $\v{w}_{\mathrm{OTLM}}$ are calculated by solving the \ref{eq:otlm} problem.

The Bayesian Mixture Models is created using skew-Gaussian functions.
The number of mixture components is the same as for the number of basis vectors in the OTLM.
Each component is a skew-Gaussian function with parameters $\mu_\mathrm{mm}, \sigma_\mathrm{mm}, \gamma_\mathrm{mm}$.
The priors are chosen to be uniform with limits set in the following way:
(i) mean $\mu_\mathrm{mm}$ is set to be $\pm 3\sigma_\mathrm{mm}$ from the X-ray energy level predicted by the \textsc{Mudirac} code for a given isotope and line,
(ii) minimum $\sigma_\mathrm{mm}$ is set to $\sigma(e)$ and maximum to $3\sigma(e)$, where $\sigma(e)$ is the energy-dependent standard deviation of the line calculated previosly for the OTLM,
(iii) minimum amplitude $w_\mathrm{mm}$ is set to the $0.1 w_{\mathrm{OTLM}}$ and maximum to $3 w_{\mathrm{OTLM}}$,
(iv) skewness $\gamma_\mathrm{mm}$ is set to $\pm 3$.
Background estimated in the pre-processing is added to the model.
Poisson likelihood is used.
The posterior is calculated using the \textsc{Emcee} sampler \citep{foreman2013emcee} with 500 walkers and 5000 steps.

\section{Appendix: Infrared spectroscopy} \label{app:IFS}
\begin{figure}
    \centering
    \includegraphics[width=0.32\linewidth, trim=0.7cm 0.3cm 1.5cm 0, clip]{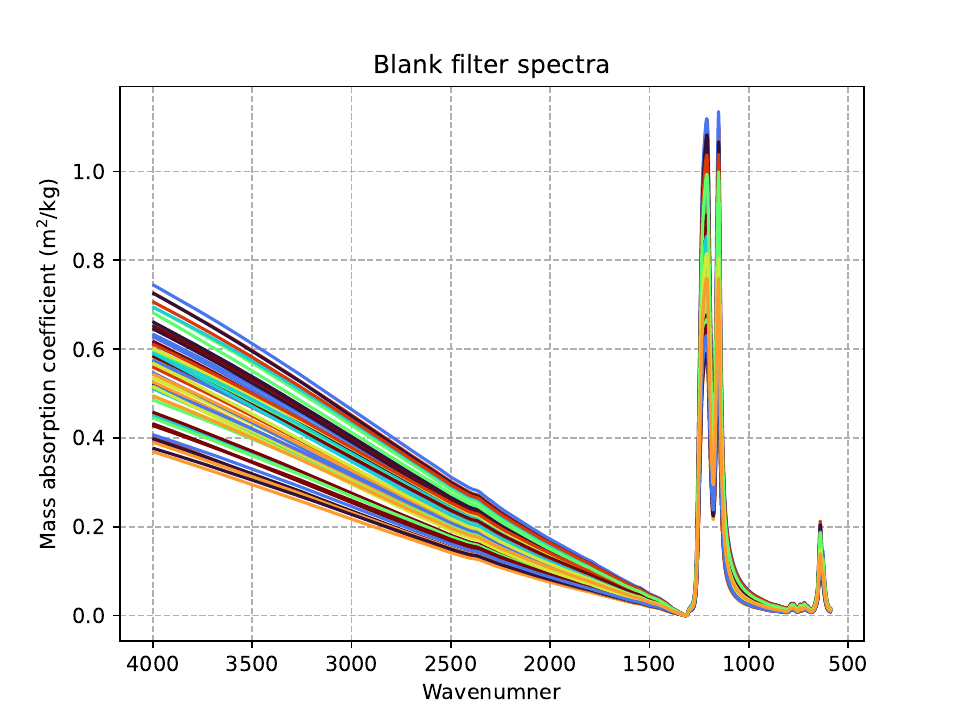}
    \includegraphics[width=0.32\linewidth, trim=0.7cm 0.3cm 1.5cm 0, clip]{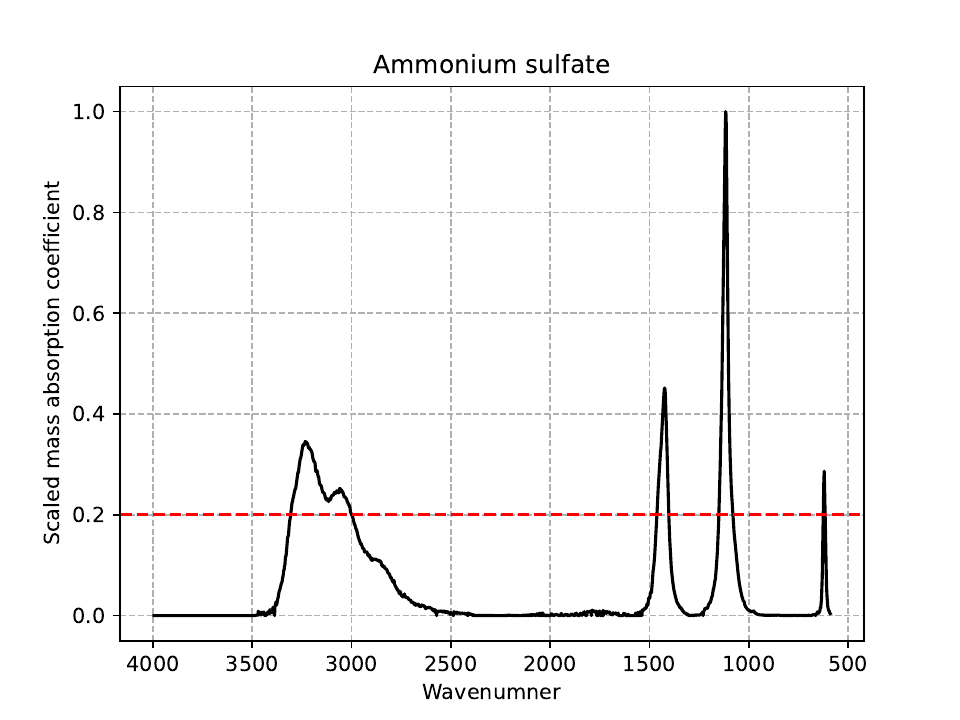}
    \includegraphics[width=0.32\linewidth, trim=0.7cm 0.3cm 1.5cm 0, clip]{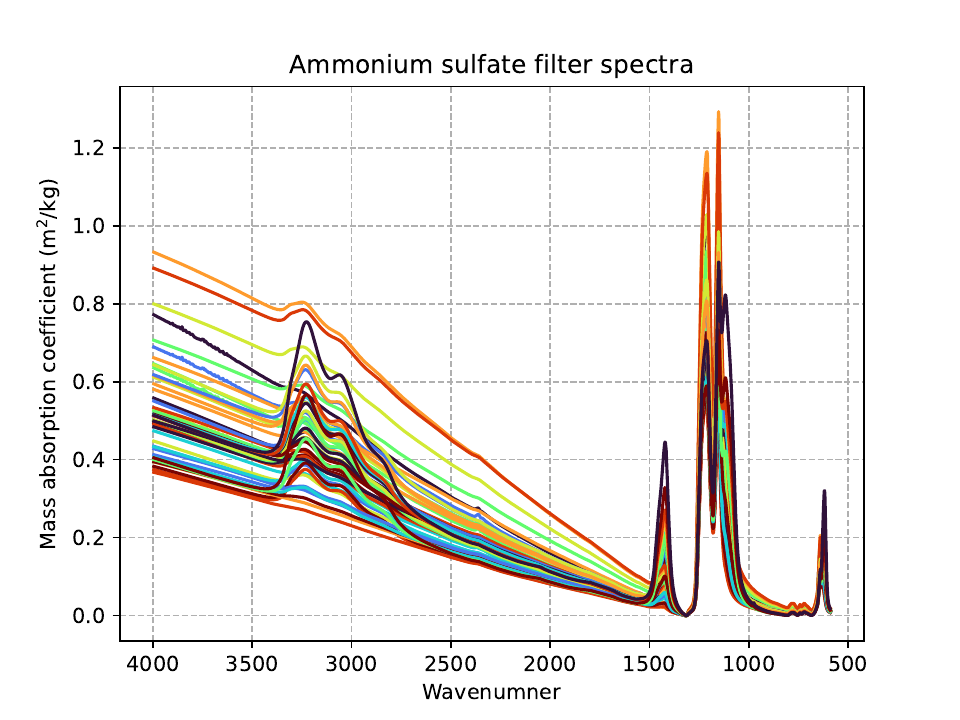}
    \caption{The 54 blank filter spectra (left panel). 
    These are used to engineer a basis for the interference. Ammonium sulfate spectra profile (middle panel) and 49 filter spectra (right panel). 
    The threshold $t$ selected for the cost matrix is shown as the red dashed line.}
    \label{fig:IRdetails}
\end{figure}

We illustrate the infrared spectroscopy data of Section \ref{sect::IRS}. We make the following remarks:
(i) All filter and reference spectra were interpoloted to wavenumbers 
$\omega\!\in\!\{500,502,...,4000\}$. and 
(ii) To make the spectra non-negative, all filters were centered to have their minimum value at zero. 
The plots show the location of the minimum value to be consistent, and therefore this preprocessing step involves minimal loss of generality.
The left panel of Figure~\ref{fig:IRdetails} contains plots of the 54 blank filter spectra used to generate a basis for the interference we can expect in the laboratory-generated analyte spectra.
The analyte spectra for ammonium sulfate is plotted in the right panels of Figure~\ref{fig:IRdetails}. 
The spectra for the other analytes are similar and not shown.
In the middle panel of this figure, the corresponding reference profile, scaled to have a maximum value of one, is displayed. 
For each analyte, the reference profile is to be added to the interference basis to reconstruct the corresponding filter spectra.
We show the thresholds $t$ of Table \ref{tab:IR_OTLM_PARAMS} as red-dashed horizontal line. 
The idea is to confine the mass in the regions above the threshold in the source marginal to the mass in the same regions in the target marginal.

\end{document}